\documentclass[runningheads]{llncs}

\usepackage{eccv}

\usepackage{eccvabbrv}

\usepackage{graphicx}
\usepackage{booktabs}
\usepackage{color}
\usepackage{tabularray}
\usepackage{tikz}
\usepackage{comment}

\usepackage[accsupp]{axessibility}  
\newcommand{\squeezeup}{\vspace{-2.5mm}}

\usepackage{hyperref}

\usepackage{orcidlink}

\begin{document}

\title{Time-Efficient and Identity-Consistent \\ Virtual Try-On Using A Variant of Altered Diffusion Models}

\titlerunning{Time-Efficient and Identity-Consistent Virtual Try-On}

\author{Phuong Dam \inst{1}\orcidlink{0009-0004-8422-3881} \and
Jihoon Jeong\inst{1}\orcidlink{0009-0003-0250-4296} \and
Anh Tran \inst{2}\orcidlink{0000-0002-3120-4036} \and 
Daeyoung Kim \inst{1}\orcidlink{0000-0002-7960-5955}}


\authorrunning{Phuong Dam, et al.}

\institute{Korea Advanced Institute of Science and Technology (KAIST) \\ \email{\{hoangphuong1211, jihooni, kimd\}@kaist.ac.kr} \and VinAI Research \\
\email{v.anhtt152@vinai.io}}


\maketitle

\begin{figure}
\centering
\scalebox{0.8}{
\includegraphics[
    width=12cm,
  keepaspectratio,
]{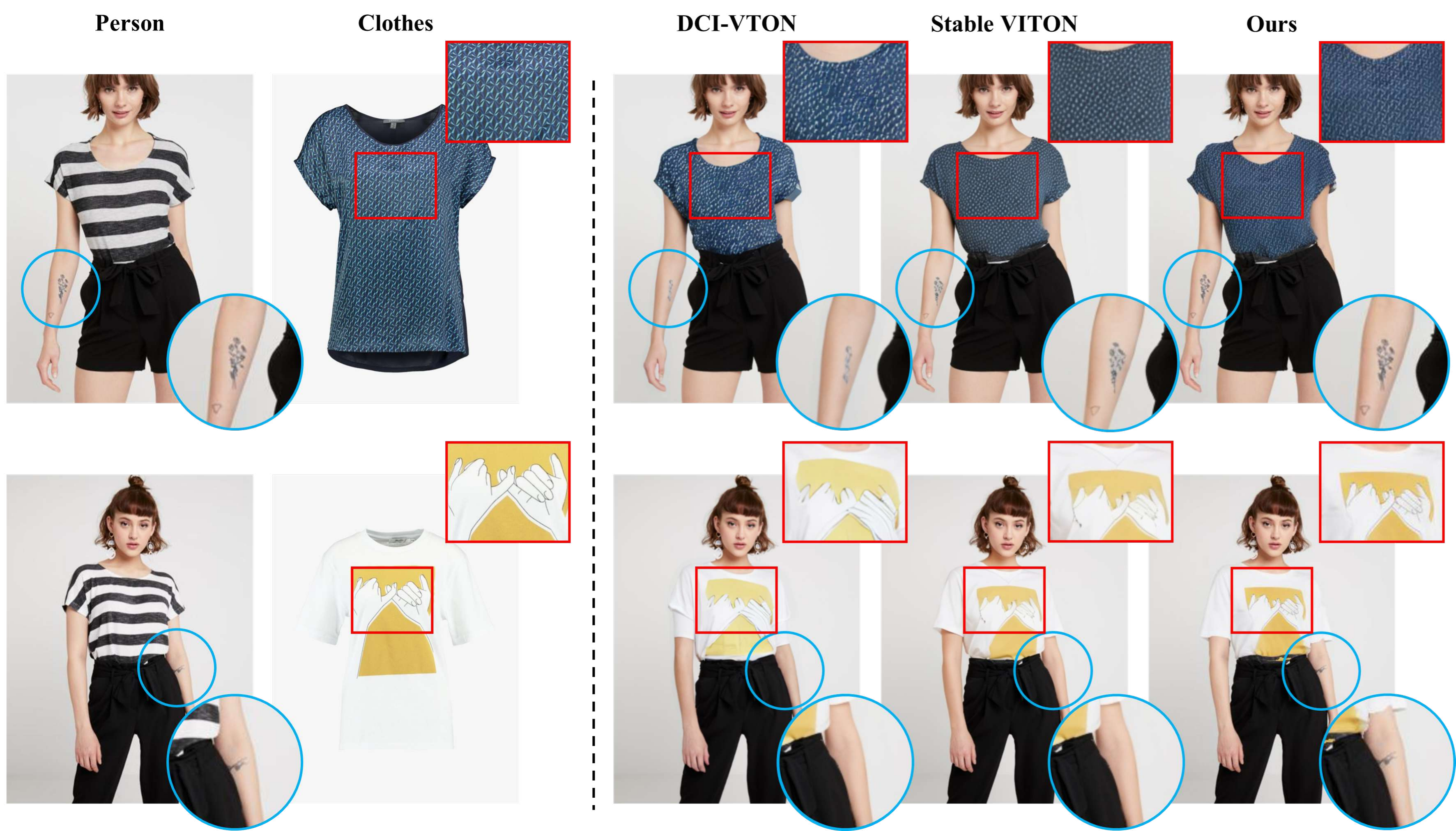}}
\caption{Visualization for identity preservation and detail preservation compared with DCI-VTON and StableVITON \cite{DCI-VTON, Stableviton}. Both models tend to degrade the texture of clothes, struggle with maintaining symbols on garments, and produce in noticeable artifacts while our approach maintains the fidelity of both garment textures and tattoos.}
\label{fig: Output-15}
\end{figure}
\squeezeup
\squeezeup
\begin{abstract}
This study discusses the critical issues of Virtual Try-On in contemporary e-commerce and the prospective metaverse, emphasizing the challenges of preserving intricate texture details and distinctive features of the target person and the clothes in various scenarios, such as clothing texture and identity characteristics like tattoos or accessories. In addition to the fidelity of the synthesized images, the efficiency of the synthesis process presents a significant hurdle. Various existing approaches are explored, highlighting the limitations and unresolved aspects, e.g., identity information omission, uncontrollable artifacts, and low synthesis speed. It then proposes a novel diffusion-based solution that addresses garment texture preservation and user identity retention during virtual try-on. The proposed network comprises two primary modules - a warping module aligning clothing with individual features and a try-on module refining the attire and generating missing parts integrated with a mask-aware post-processing technique ensuring the integrity of the individual's identity. It demonstrates impressive results, surpassing the state-of-the-art in speed by nearly 20 times during inference, with superior fidelity in qualitative assessments. Quantitative evaluations confirm comparable performance with the recent SOTA method on the VITON-HD and Dresscode datasets. We named our model \textbf{F}ast and \textbf{I}dentity \textbf{P}reservation \textbf{Vi}rtual \textbf{T}ry\textbf{ON} (FIP-VITON).

\end{abstract}

\section{Introduction}
\label{sec:intro}

Virtual Try-On, which involves placing a garment on a particular individual, holds crucial significance in contemporary e-commerce and the prospective metaverse. The key challenge lies in preserving intricate texture details and distinctive features of the target person, such as appearance and pose. Adapting a garment to different body shapes without altering patterns is particularly challenging, especially when body pose or shape varies significantly. 

Recent studies based on deep learning techniques have approached these challenges by defining specific body-garment corresponding regions, particularly addressing obstructions \cite{GP-VTON} or by adding cloth segmentation information \cite{HR-VTON}. Another approach is taking advantage of the strength of the Diffusion network, combined with Contrastive Language-Image Pretraining (CLIP) \cite{CLIP}, which refines post-warping clothing results along with generating missing body parts \cite{DCI-VTON}. Another method uses implicit warping integrated with Diffusion, guided by CLIP-based networks, to overcome this limitation \cite{TryonDiffusion}. All the diffusion-based approaches are proven to surpass the traditional flow-based methods in both quantitative and qualitative assessment \cite{DCI-VTON,TryonDiffusion}.

Despite strong generative ability, diffusion-based approaches suffer from extended inference times and uncontrollable artifact generation, affecting the user experience and image fidelity. Meanwhile, another equally important challenge besides the garment texture preservation is retaining the user's identifying characteristics during virtual try-on - mentioned in \cite{TryonDiffusion}. Therefore, to tackle these issues, we propose a novel diffusion approach that not only effectively preserves both the garment texture and identity information but also achieves an impressive inference speed for this task.

Our network comprises two primary modules - a warping module and a try-on module, integrated with post-processing blocks. The warping module is pivotal in aligning clothing with the individual's features. It considers clothing specifics and person-related information, encompassing key points, dense pose images, and garment type-specific regions of interest (e.g., upper, lower, or full dresses). 
Subsequently, the try-on module refines the warped attire from the initial module, generating the missing parts of the image. The image then undergoes a conditional post-processing named mask-awareness technique to ensure the fundamental integrity of the individual's identity. Examples of our impressive results compared to those from the SOTA papers \cite{DCI-VTON,Stableviton} are depicted in detail in Fig. \ref{fig: Output-15}.

In summary, the main contributions of our work are:
\begin{itemize}
    
    \item We introduce a novel try-on technique to generate photo-realistic results for diverse scenarios.  
    
    \item We introduce a novel time-efficient diffusion approach that can adjust and maintain the garment details and generate the missing body parts using sophisticated conditional modules, which effectively guide the model's focus during the generation process to yield satisfying outcomes. 

    \item We introduce a mask-aware post-processing technique that not only preserves the individual's identity details but also improves the overall fidelity of the generated images.
    
\end{itemize}

\section{Related Work}
\subsection{Virtual Try-on GAN-based Models}
In virtual try-on, achieving high levels of realism and fidelity in garment rendering on digital avatars remains a significant challenge. Traditional deep-learning approaches, primarily utilizing flow-based Generative Adversarial Networks \\ (GANs), have demonstrated notable potential. Most existing virtual try-on methods follow a two-stage process \cite{HR-VTON,VITON-HD,GP-VTON}. The first stage involves a warping module responsible for predicting the appearance flow for the global \cite{HR-VTON,VITON-HD}, or local parts \cite{GP-VTON} of the garment to fit the target person's pose. This is followed by a second stage, where a GAN-based generator seamlessly integrates the warped garment into the model. Although this method has shown some effectiveness, its heavy reliance on warping quality in the initial stage often leads to less-than-ideal outcomes. This is particularly evident in the realism of garment-skin boundaries and the overall try-on effect in the garment area. 

Despite their widespread use, these methods have not seen significant innovation to overcome these specific challenges. \emph{Therefore, we proposed an approach that aims to push the boundaries of the field by investigating the use of diffusion-based generators.}
\squeezeup
\subsection{Virtual Try-on Diffusion Models}
Diffusion models have recently emerged as formidable rivals to GANs in image generation, excelling in producing high-fidelity conditional images. Their application ``diffuses'' a diverse range of tasks, including text-to-image generation, image-to-image translation, and image editing. Notably, in the context of virtual try-on, diffusion models offer a promising solution by treating the task as a form of image editing conditioned on a specific garment and a full-body image.

A notable diffusion-based approach is presented in TryonDiffusion \cite{TryonDiffusion}, which leverages a cross-attention mechanism and integrates an implicit warping algorithm with a try-on module. While this method showcases potential in virtual try-on applications, \emph{it struggles to retain textural details in the final output.} 

Other strategies combine the explicit warping module from GAN-based methods with a diffusion model to merge warped clothing and person images. LaDI-VTON \cite{LaDI-VTON}, DCI-VTON \cite{DCI-VTON}, and StableVITON \cite{Stableviton} leverage the capabilities of Stable Diffusion to preserve the texture and details of in-shop garments, achieving high-quality images to a certain level. However, these methods often create unintended artifacts in the final images, which remain challenging to control. Despite the outstanding generation quality of the diffusion approach, \emph{the models' inference times are excessively long}. The major reason lies in the  \emph{large number of denoising steps during the inference process.} 
\squeezeup
\subsection{Diffusion Model Speed Up Techniques}
Recent research has focused on accelerating the inference time of Diffusion model networks. Most of these methods are based on types of distillation techniques, such as \cite{consistency,lcm,distillation}. Meanwhile, in specific terms of virtual try-on, no generalized approaches for multiple timestep diffusion have been distilled. Although \cite{consistency} offers a direct training method that allows generating images with a single timestep, there is insufficient evidence that this method works well in a conditional high-resolution diffusion model, leaving space for future research. Furthermore, several studies applied a multimodal conditional GAN to reduce the number of timesteps to 4 or 2 \cite{DDGAN,WaveDiff}, speeding up the inference process of the diffusion model. \emph{Inspired by leveraging multimodal GAN} \cite{DDGAN,WaveDiff}, we opt to experiment with a single-step diffusion-based approach employing a fixed noise level based on Denoising Diffusion Implicit Models (DDIM) \cite{DDIM}.
\section{Methodology}
Addressing these challenges, we propose a novel approach to enhance artifact control. Our method involves developing a diffusion model that not only competes with state-of-the-art(SOTA) models' performance but also ensures time efficiency. This is achieved by incorporating \emph{(un)conditional mask-aware techniques} and \emph{a modified adding noise algorithm reducing the number of time steps to one}, thereby offering a practical and effective solution for virtual try-on applications. The mask-aware techniques are particularly crucial in attaining a realistic wearing effect, as they directly address the common issue of unnatural transitions between clothing and skin, and preserve individual’s identities. 

\begin{figure}[t]
\centering
\scalebox{0.9}{\includegraphics[
  width=12cm,
  height=6cm,
  keepaspectratio,
]{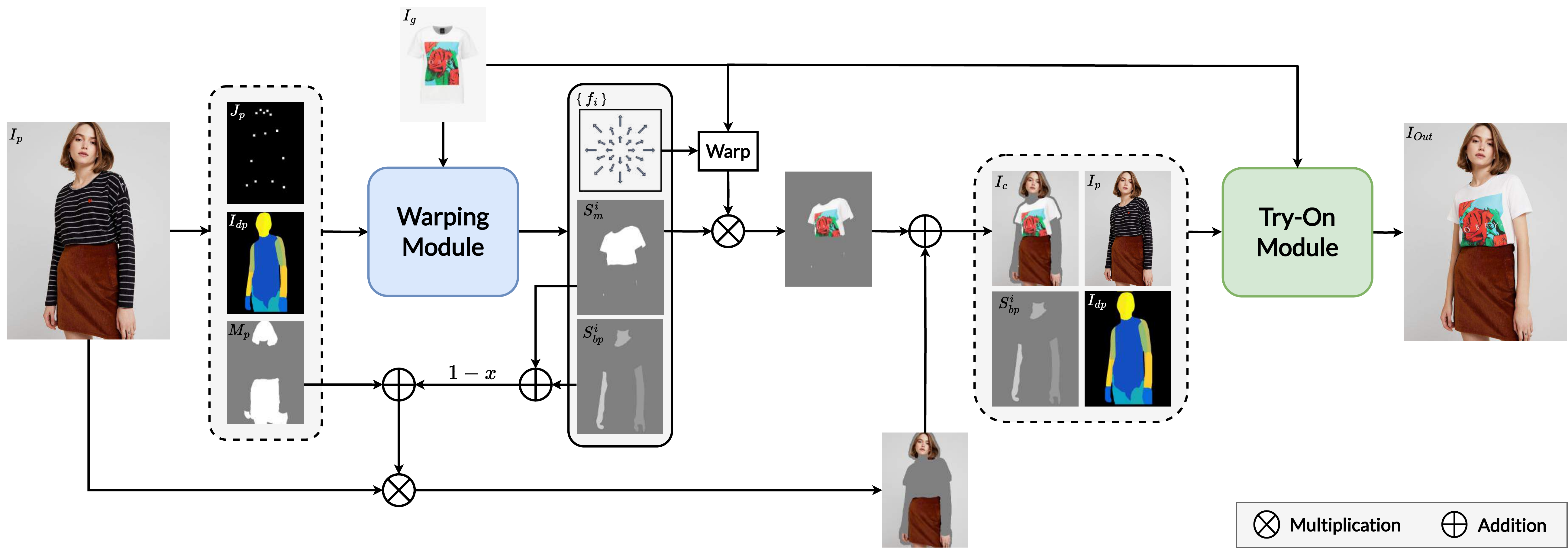}}
\caption{Overall Generation Pipeline.}
\label{fig:full_flow_pipeline}
\end{figure}

In the virtual try-on task, given an image of a person \(I_p\) and an image of a garment \(I_g\), we want to obtain the try-on image that portrays the person wearing the garment. The overall architecture of our approach is depicted in Figure \ref{fig:full_flow_pipeline}. We first pre-process the person image to obtain reference information, including human parsing, 2D pose key points \(J_p\), and dense pose \(I_{dp}\). Following this, our wrapping module utilizes the reference information and produces predicted multi-scale warped-based parsing \{\(S_i\)\}\ including both the warped cloth mask \(S^i_m\) and the segmentation of visible body parts \(S^i_{bp}\), along with multi-scale appearance flows \{\(f_i\)\}. 
At the highest resolution (\(i=1\)), these outputs are integrated with the garment and person images, in addition to preserved person information (\(M_p\)), to construct the conditional input (\(I_c\)). The conditional input \(I_c\) in conjunction with dense pose \(I_{dp}\), predicted body part segmentation \(S^i_{bp}\), garment image \(I_g\), and person image \(I_p\) are fed into the try-on module to produce the final image \(I_{Out}\).
\squeezeup
\subsection{Preprocessing}
The procedure begins with generating human parsing maps from the person image (\(I_p\)) by employing advanced human parsing methods \cite{HumanParsing}. Then, we apply 2D key points \cite{OpenPose}, and dense pose estimation \cite{DensePose}; their outputs are denoted as \(J_p\) and \(I_{dp}\), respectively. Subsequently, depending on the garment type, we \emph{identify and exclude mutable sections} based on the human parsing map. Specifically, this mechanism is that if the garment is upper types, the upper regions of the body are omitted, which works the same as the lower types and dresses – the full body types. The remaining segments are combined to create the preserved mask (\(M_p\)).  
\squeezeup
\begin{figure}[t]
\centering
\scalebox{0.8}{
\includegraphics[
  width=12cm,
  keepaspectratio,
]{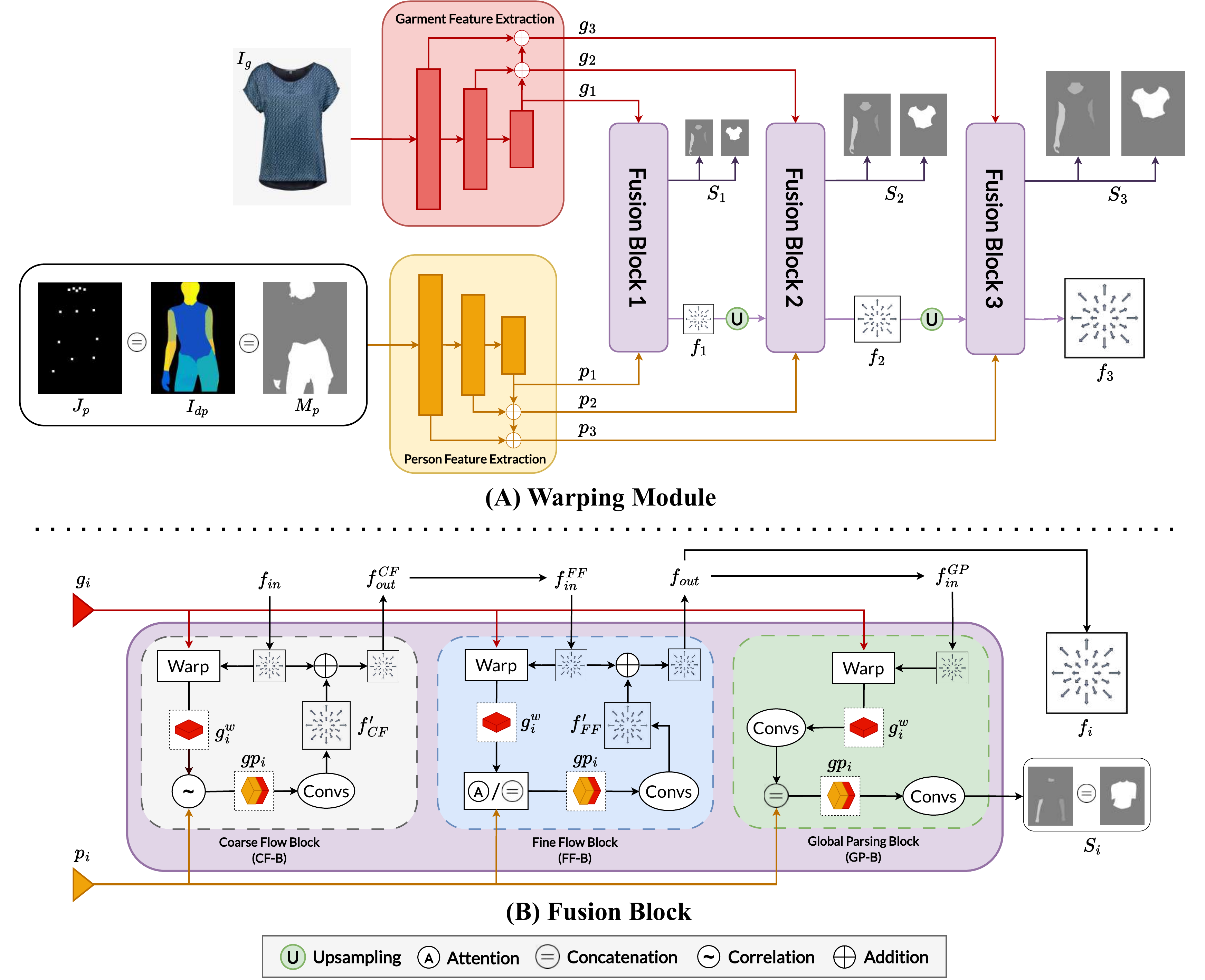}}
\caption{\textbf{Warping Module structure}. It is crucial to highlight that our model extracts six or seven multi-scale features, depending on the input resolution (i.e., $N$ = 6 or 7). For brevity, the number of scales depicted in this figure is limited to three ($N$ = 3).}
\label{fig: warping module}
\end{figure}
\squeezeup
\subsection{Warping Module}
In this section, we detail the structure of the Warping Module. As illustrated in Figure \ref{fig: warping module}, our warping module draws inspiration from the flow estimation pipeline found in \cite{HR-VTON,Style-based,Clothflow,Parser-free,GP-VTON}. It comprises two pyramid feature extraction blocks, "Garment Feature Extraction" and "Person Feature Extraction" modules; cascade flow estimation. Our distinct contributions in this module will be elucidated below.
\squeezeup
\subsubsection{Pyramid Feature Extraction.}
Our warping module leverages two Feature Pyramid Networks (FPN) \cite{FPN} for extracting $N$ multi-scale person and garment features. The person feature extraction block receives inputs from the 2D human pose keypoints map \(J_p\), dense pose \(I_{dp}\), and the preserving region mask \(M_p\). Conversely, the garment feature extraction block exclusively takes inputs from the intact in-shop garment \(I_g\). Notably, we have no change compared to the FPN \cite{FPN} used in \cite{GP-VTON} except the number of scales ($N$) varies on input resolution. 
\squeezeup
\subsubsection{Cascade Flow Estimation.}
Inspired by established methods \cite{HR-VTON,Style-based,Clothflow,Parser-free,GP-VTON}, instead of estimating the local flows for certain parts of the cloth as in \cite{GP-VTON}, we target the direct prediction of global flow for warping intact garments. Meanwhile, our warping module, depicted in Figure \ref{fig: warping module}, adopts concepts from \cite{GP-VTON} with internal enhancement. \emph{Notably, our enhancement introduces cross-attention for additional feature integration, elevating the quality of the warping outcome and thereby enhancing the overall model performance.} 

Specifically, our module incorporates $N$ Fusion blocks designed to handle multi-scale flow maps and human parsing predictions. Each Fusion block is composed of a Coarse Flow Block (CF-B), a Fine Flow Block (FF-B), and a Global Parsing Block (GP-B), depicted in gray, blue, and green, respectively, Fig. \ref{fig: warping module}(B). 

In CF-B, the garment feature \(g_i\) undergoes warping with incoming flow \(f_{in}\) to produce \(g^w_i\). The correlation operation from FlowNet2 \cite{Flownet} integrates it with the person feature \(p_i\), and subsequent convolution layers estimate the corresponding flow \(f'_{CF}\). The refined coarse flows \(f^{CF}_{out}\) result from adding \(f'_{CF}\) to \(f_{in}\). FF-B, sharing CF-B's architecture, treats \(f^{CF}_{out}\) as the input flow \(f^{FF}_{in}\). Diverging from CF-B, FF-B opts for multi-head cross-attention instead of correlation, using scaled dot-product attention \cite{Attention}:
\begin{equation}
\centering
\textrm{Attention} (Q, K, V) = \textrm{softmax}(\frac{QK^T}{\sqrt{d}})V\
\label{equa:attn}
\end{equation}
where \begin{math} Q \in \mathbb{R}^{M \times d}, K \in \mathbb{R}^{N \times d}, V \in \mathbb{R}^{N \times d} \end{math} represent stacked vectors of query, key, and value, respectively. \(M\) is the number of query vectors, \(N\) is the number of key and value vectors, and \(d\) is the dimension of the vector. In our setup, \(Q\) represents the flattened feature of the warped garment \(g^w_i\), while \(K\) and \(V\) correspond to the flattened features of the person \(p_i\). The dot-product-based attention map \begin{math} \frac{QK^T}{\sqrt{d}}\ \end{math} serves as an additional feature indicating the similarity between the person and the warped garment. Moreover, we restrict the application of multi-head cross attention to the feature resolution below 64x48 for parameter efficiency, switching to concatenation at larger resolutions. The output of this operation is directed to a group convolution block for estimating flow \(f'_{FF}\), which is added to \(f^{FF}_{in}\) to yield the fine flow \(f_{out}\).

Within the Global Parsing Block (GP-B), leveraging the enhanced flow \(f^{GP}_{in}\), the garment feature \(g_i\) is warped. The newly warped feature \(g^w_i\) undergoes fusion with the incoming person feature \(p_i\) through convolution operations. The concatenated feature \({gp}_i\) is then processed by convolutional layers to estimate the global parsing \(S_i\), covering background, cloth, left/right arms, center body parts (including neck and belly), and left/right legs.
\squeezeup
\subsubsection{Objective Function.}
Similar to numerous prior studies employing flow-based warping models \cite{HR-VTON,Style-based,Clothflow,Parser-free,GP-VTON}, our warping model follows a similar structure, incorporating a combination of various loss functions. In the training of the warping module, we utilized $\ell_1$ loss - \(L^w_1\) and perceptual loss \cite{PerceptualLoss} - \(L^w_{per}\) for the warped result. Additionally, pixel-wise cross-entropy loss \(L_{ce}\), and  $\ell_1$ loss \(L^w_m\) are applied to the entire estimated parsing. Adversarial loss \cite{RelativisticGAN} - \(L^w_{adv}\) is employed for both overall parsing and the warped result. Given the appearance flow's high degree of freedom, we also include total-variation (TV) loss \(L_{TV}\) as proposed in \cite{DCI-VTON}, effectively addressing the smoothness of the final warping result. In alignment with the approach outlined in \cite{Parser-free}, we augment a second-order smooth constraint \(L_{sec}\). The total loss function for our warping module can be formulated as:
\begin{equation}
\centering
L^w = L^W_1 + \alpha^w_{per}L^w_{per} + \alpha_{ce} L_{ce} + \alpha^w_m L^w_m + \alpha^w_{adv} L^w_{adv} + \alpha_{TV}L_{TV} + \alpha_{sec}L_{sec},
\label{equa:warpingmodule_loss}
\end{equation}
with \(\alpha^w_{per}\), \(\alpha_{ce}\), \(\alpha^w_m\), \(\alpha^w_{adv}\), \(\alpha_{TV}\), and \(\alpha_{sec}\) are hyper-parameters controlling the contribution of each loss term to the overall loss.
\squeezeup
\begin{figure}[t]
\centering
\scalebox{0.8}{
\includegraphics[
  width=12cm,
  keepaspectratio,
]{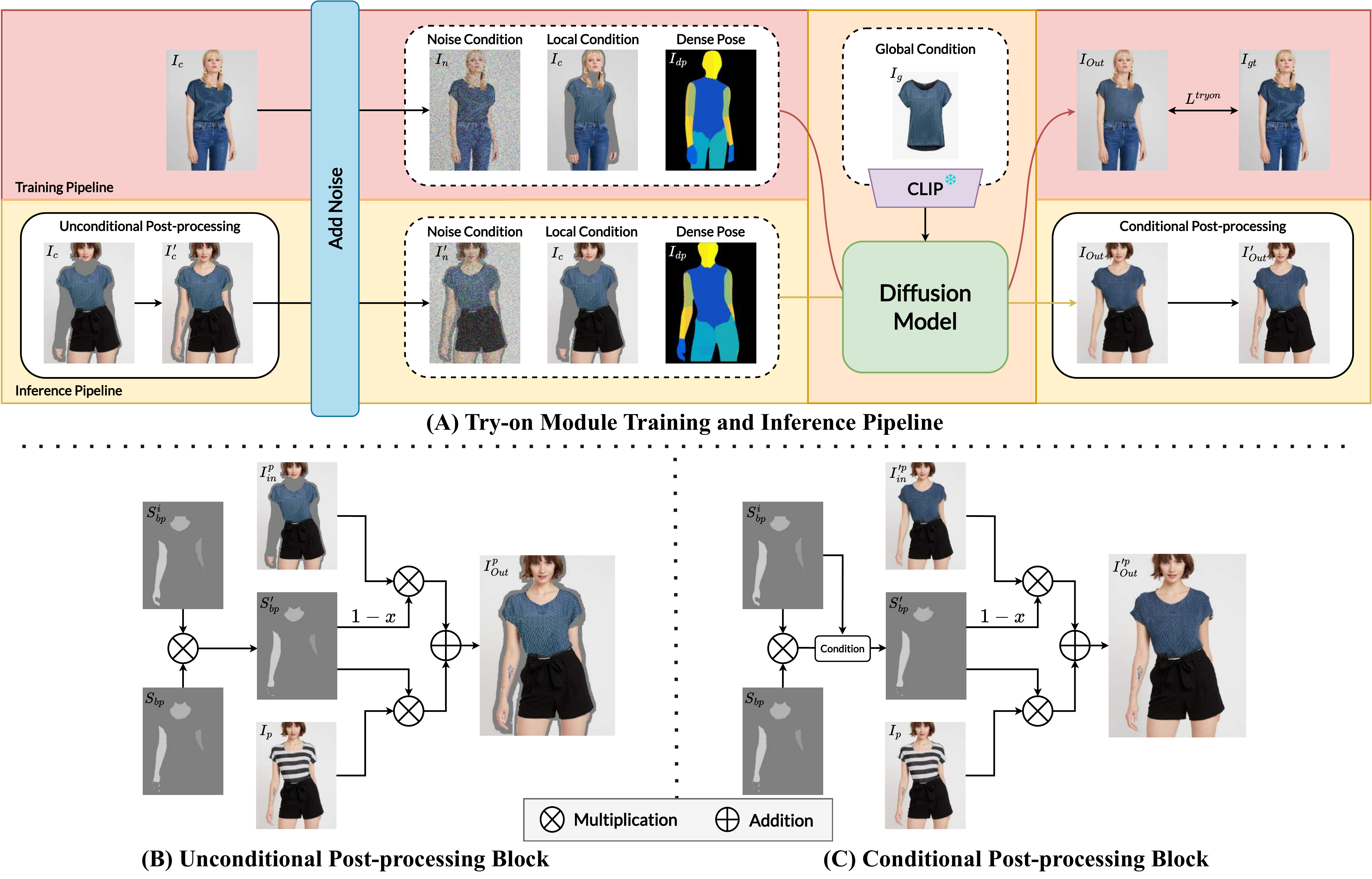}}
\caption{Try-on module training, inference strategy, and details. \(S_{bp}\) is the body part segmentation extracted from \(I_p\).}
\label{fig: Try-on Training Pipeline}
\end{figure}

\subsection{Try-on Module}
Despite the robust stability exhibited by traditional vanilla diffusion models, they suffer from a notable drawback — prolonged inference times, which can hinder practical applications. \emph{However, our variance approach addresses this issue without compromising the model's generative performance.} As indicated in the overview of our strategy in Fig.\ref{fig:full_flow_pipeline}, we intend to apply a variant of the modified diffusion model to refine the coarse synthesis result. Following the acquisition of the warping and parsing result, the warped garment is incorporated into the person's image, alongside the preserved elements and the original background, to form the conditional image \(I_c\). This image \(I_c\) serves as the local condition for the try-on diffusion module. Moreover, the global conditions are derived from the target garment image \(I_g\) using the feature representations extracted by a frozen, pre-trained CLIP \cite{CLIP} image encoder. All are illustrated in Fig. \ref{fig: Try-on Training Pipeline}(A). 
\squeezeup
\subsubsection{Training pipeline.}
During the training process, we work with pair-setting; a noise level is directly added to the ground truth image \(I_{gt}\) to create a conditional noise input \(I_n\). Subsequently, the local condition image \(I_c\) mentioned earlier is concatenated with the dense pose image \(I_{dp}\) - dense pose condition, and the conditional noise \(I_n\). This concatenated input is then fed into the model. The output from the model is then merged with the static components, which include the overlapping areas of the predicted and actual backgrounds as well as the unaltered segment \(M_p\). \emph{This highlights our try-on model's focus on generating missing image parts and refining clothing details.} 

The objective function is then applied to the integrated output and the ground truth image. Our model leverages global and local conditions in the optimization process to generate the corresponding inferred person image. In contrast to prior studies related to the diffusion model, where the amount of noise added to the image is adjusted through a time variable, in this study, we use a fixed amount of noise, rendering the time variable redundant during the training process. \emph{At this point, our approach could be considered a single-step diffusion model.} The adding noise process follows the equation:
\begin{equation}
\centering
z = z_0 + \alpha_n \epsilon,
\end{equation}
where \(\alpha_n\) is the noise ratio and \(\epsilon \sim \mathcal{N}(0,I)\). Therefore, our framework is built upon the baseline model of the DDIM study \cite{DDIM}, with the temporal embedding block supplanted by the CLIP feature.

\subsubsection{Inference pipeline.}
In the absence of the ground truth image during inference, it becomes imperative to generate an alternative conditional input to facilitate the inference pipeline. In the inference process, the local condition \(I_c\) is constructed and sent to a post-processing block after the warping module. Specifically, the output of the first post-processing block \(I'_c\) is created by combining the local condition image \(I_c\) with the remaining body parts extracted from the ground truth image \(I_p\). This combined image is then made noisy to obtain another noise-conditioned image \(I'_n\).  Following a similar diffusion process as outlined in the training phase, the output image \(I_{Out}\) is advanced into another post-processing block, culminating in the final result \(I'_{Out}\). This sequence ensures the generation of a refined output image, compensating for the absence of ground truth data through strategic conditioning and processing stages.
\squeezeup
\subsubsection{(Un)conditional Post-processing Block.}
The post-processing technique, termed (un)conditional mask-aware, operates throughout the entire inference pipeline, ensuring the enhancement of synthesized images with a focus on artifact reduction and detail preservation. \emph{The first post-processing technique, unconditional post-processing, aims to enhance the condition image, improving the final synthesized results. Meanwhile, the second technique, conditional post-processing, focuses on minimizing artifacts and preserving identity details.}

The unconditional post-processing block involves obtaining body segmentation masks \(S^i_{bp}\) here $i = 1$ denoting the largest segmentation map extracted from the warping module and corresponding body masks \(S_{bp}\) from the reference person image. The overlapping mask \(S'_{bp}\) is used to extract unchanged elements from \(I_p\), which are then added to \(I_c\) to get the coarse try-on image \(I'_c\).

The conditional post-processing block is used to re-apply the unchanged parts from the input image \(I_p\) to the refined try-on image $I_{Out}$, correcting any undesirable modifications caused by the diffusion model. It has the same structure as the unconditional one, except that it is only applied when the overlapping ratio between \(S_{bp}\) and \(S^i_{bp}\) where $i = 1$, is greater than a threshold. Empirically, we found this threshold $R_{\textrm{overlap}}$ needs to be greater than 0.8. This experiment is detailed in the Appendix, providing the applied rate of each specific threshold. 
\begin{equation}
\centering
R_{\textrm{overlap}} = \frac{\left | S^{(1)}_{bp} \bigcap  S_{bp}\right |}{S^{(1)}_{bp}} = \frac{S'_{bp}}{S^{(1)}_{bp}} 
\label{equa:mask_iou}
\end{equation}
Notably, we apply this to every single parsing part, including the left/right hand, left/right leg, and center body part (neck and belly), shown in Fig. \ref{fig: Try-on Training Pipeline}(B),(C). 
\squeezeup    
\subsubsection{Objective function.} In terms of optimization processing for the try-on module, we utilize 2 reconstruction losses which are $\ell_1$ loss - \(L^{\textrm{tryon}}_1\) and VGG perceptual loss \cite{PerceptualLoss} - \(L^{\textrm{tryon}}_{per}\). Furthermore, an adversarial loss from \cite{RelativisticGAN} is also used for better quality results - \(L^{\textrm{tryon}}_{adv}\). The total loss for the try-on module can be formulated:
    
\begin{equation}
\centering
L^{\textrm{tryon}} = L^{\textrm{tryon}}_1 + \alpha^{\textrm{tryon}}_{per}L^{\textrm{tryon}}_{per} + \alpha^{\textrm{tryon}}_{adv} L^{\textrm{tryon}}_{adv}
\label{equa:tryon_loss}
\end{equation}    
 where, \(\alpha^{\textrm{tryon}}_{per}\) and \(\alpha^{\textrm{tryon}}_{adv}\) are the balance coefficients.

\section{Experiments}
\subsection{Experiments Setting}
\subsubsection{Dataset.}
Our experiments primarily use the VITON-HD dataset \cite{VITON-HD}, comprising 13,679 frontal-view woman and upper clothes image pairs at 1024x768 resolution. Following prior work \cite{VITON-HD,HR-VTON}, we split the dataset into a training set of 11,647 pairs and a test set of 2,032 pairs. Experiments are conducted at various resolutions, and we also assess the model on the DressCode dataset \cite{DressCode} for added complexity. Much like VITON-HD \cite{VITON-HD}, the DressCode dataset \cite{DressCode} is a repository of high-quality try-on data pairs, comprising three distinct sub-datasets: dresses, upper-body, and lower-body. In total, the dataset encompasses 53,795 image pairs, distributed across 15,366 pairs for upper-body attire, 8,951 pairs for lower-body clothing, and 29,478 pairs for dresses.
To maintain consistency, we apply the same human parsing and key points pose estimation methods used on the DressCode \cite{DressCode} to the VITON-HD \cite{VITON-HD}.
\subsubsection{Evaluation Metrics.}
In Virtual Tryon evaluations, we consider paired and unpaired settings. Paired assessments use Structural Similarity Index Measure (SSIM) \cite{SSIM} and Learned Perceptual Image Patch Similarity (LPIPS) \cite{LPIPS} for image reconstruction, while unpaired settings employ Frechet Inception Distance (FID) \cite{FID} and Kernel Inception Distance (KID) \cite{KID} to measure the model's ability to generate new images with changed clothing. Specifically, our evaluation metrics for VITON-HD \cite{VITON-HD} are all above. Meanwhile, all experiments conducted on the DressCode dataset \cite{DressCode} are executed at a resolution of 512 × 384, and the evaluation metrics only include LPIPS \cite{LPIPS}, SSIM \cite{SSIM}, and FID \cite{FID}. In addition, we also measure the speed of synthesizing at the 512 x 384 resolution (T(s)) of the VITON-HD dataset. For a fair comparison in terms of inference time, we ensure consistent configuration settings, employing a single RTX 4090 with a batch size of 4. The model's inference time is computed by averaging the time taken for end-to-end inference over the entire testing set, repeated 10 times.
\squeezeup
\subsection{Quantitative Evaluation}
In our comparative analysis with existing virtual try-on methods on VITON HD dataset \cite{VITON-HD} and DressCode dataset \cite{DressCode}, including CP-VTON \cite{CP-VTON}, VITON-HD \cite{VITON-HD}, FS-VTON \cite{Style-based}, SDAFN \cite{SDAFN}, PF-AFN \cite{Parser-free}, HR-VTON \cite{HR-VTON}, GP-VTON \cite{GP-VTON}, LaDI-VITON \cite{LaDI-VTON}, StableVITON \cite{Stableviton}, and the current state-of-the-art (SOTA) DCI-VTON \cite{DCI-VTON}, our method showcases competitive performance across various metrics (Table \ref{tab:Quantitative_Table}). In terms of VITON HD \cite{VITON-HD}, our model demonstrates compatibility with this SOTA method while DCI-VTON outperforms previous studies in all evaluated categories. Notably, our approach excels in certain aspects while trailing in others, both in paired and unpaired settings. A significant advantage of our model lies in its remarkable inference speed. As detailed in Table \ref{tab:Quantitative_Table}, our model achieves the best inference times, which is more than 17.43 times faster than DCI-VTON in 512x384 resolution, 1.01s of ours compared to 17.60s of DCI-VTON. Besides, our method also demonstrates superior performance that is compatible with the DCI-VTON in almost all three subsets of DressCode \cite{DressCode}.
\squeezeup
\begin{table}[b]
\centering
\caption{\small{Quantitative comparison with baselines on VITON-HD \cite{VITON-HD} and DressCode \cite{DressCode}. We multiply KID by 100 for better comparison. T(s)↓ is the inference times. \textbf{Bold} and \underline{underline} denote the best and the second best.}}
\scalebox{0.44}{
\begin{tblr}{
  cells = {c},
  cell{1}{1} = {r=3}{},
  cell{1}{2} = {c=9}{},
  cell{1}{11} = {c=9}{},
  cell{2}{2} = {c=4}{},
  cell{2}{6} = {c=5}{},
  cell{2}{11} = {c=3}{},
  cell{2}{14} = {c=3}{},
  cell{2}{17} = {c=3}{},
  vline{2,11} = {1}{},
  vline{2,6,11,14,17} = {2}{},
  vline{2,6,11,14,17} = {3}{},
  vline{2,6,11,14,17} = {4-14}{},
  hline{1,4,11,15} = {-}{},
  hline{2-4} = {2-19}{},
}
\textbf{Method} & \textbf{VITON-HD}  &                &               &                &                    &                &               &                &               & \textbf{DressCode - (512 x 384)} &                 &               &                 &                 &                &                 &                 &               \\
                & 256 x 192 &                &               &                & 512 x 384 &                &               &                &               & Upper                   &                 &               & Lower  &                 &                & Dress  &                 &               \\
                & LPIPS↓    & SSIM↑ & FID↓ & KID↓  & LPIPS↓    & SSIM↑ & FID↓ & KID↓  & T(s)↓ & LPIPS↓                  & SSIM↑  & FID↓ & LPIPS↓ & SSIM↑  & FID↓  & LPIPS↓ & SSIM↑  & FID↓ \\
CP-VTON \cite{CP-VTON}         & 0.089              & 0.739          & 30.11         & 2.034          & 0.141              & 0.791          & 30.25         & 4.012          & -             & -                                & -               & -             & -               & -               & -              & -               & -               & -             \\
VITON-HD \cite{VITON-HD}        & 0.084              & 0.811          & 16.36         & 0.871          & 0.076              & 0.843          & 11.64         & 0.3            & \textbf{0.64}             & -                                & -               & -             & -               & -               & -              & -               & -               & -             \\
FS-VTON \cite{Style-based}        & -                  & -              & -             & -              & -                  & -              & -             & -              & -             & 0.0376                           & 0.9457          & 13.16         & 0.0438          & 0.9381          & 17.99          & 0.0745          & \textbf{0.8876} & 13.87         \\
SDAFN \cite{SDAFN}          & -                  & -              & -             & -              & -                  & -              & -             & -              & -             & 0.0484                           & 0.9379          & 12.61         & 0.0549          & 0.9317          & 16.05          & 0.0852          & 0.8776          & 11.8          \\
PF-AFN \cite{Parser-free}          & 0.089              & 0.863          & 11.49         & 0.319          & 0.082              & 0.858          & 11.3          & 0.283          & -             & 0.038                            & 0.9454          & 14.32         & 0.0445          & 0.9378          & 18.32          & 0.0758          & 0.8869          & 13.59         \\
HR-VTON \cite{HR-VTON}        & 0.062              & 0.864          & 9.38          & 0.153          & \underline{0.061}              & 0.878          & 9.9           & 0.188          & 1.39  & 0.0635                           & 0.9252          & 16.86         & 0.811           & 0.9119          & 22.81          & 0.1132          & 0.8642          & 16.12         \\
GP-VTON \cite{GP-VTON}        & -                  & -              & -             & -              & 0.08               & 0.894          & 9.2           & -              & -             & 0.0359                           & \underline{0.9479}  & 11.89         & 0.042           & \underline{0.9405}          & 16.07          & 0.0729          & \underline{0.8866}  & 12.26         \\

LaDI-VITON \cite{LaDI-VTON}       & -     & -  & -  & - & 0.0986     & 0.858  & 12.31 & 0.567 & 8.27        & 0.0654                  & 0.9129               & 16.18 & 0.0603 & 0.9076       & 16.31 & 0.1079 & 0.852               & 15.80 \\

StableVITON \cite{Stableviton}      & -     & -  & -  & - & 0.073     & 0.888  & 8.58 & 0.073 & 20.58        & 0.0388                  & 0.937               & \textbf{9.94} & - & -               & - & - & -               & - \\

DCI-VTON \cite{DCI-VTON}       & \textbf{0.049}     & \underline{0.906}  & \underline{8.02}  & \textbf{0.058} & \textbf{0.043}     & \underline{0.896}  & \textbf{8.09} & \textbf{0.028} & 17.60        & \textbf{0.0301}                  & -               & 10.82 & \textbf{0.0348} & -               & \textbf{12.41} & \textbf{0.0681} & -               & \underline{12.25} \\

FIP-VITON (Ours)            & \underline{0.056}      & \textbf{0.909} & \textbf{7.53} & \underline{0.07}   & 0.067       & \textbf{0.909} & \underline{8.43}  & \underline{0.066}  & \underline{1.01} & \underline{0.0357}                   & \textbf{0.9495} & \underline{10.4} & \underline{0.0417}  & \textbf{0.9413} & \underline{12.69}  & \underline{0.0727}  & 0.886           & \textbf{11.2} 
\end{tblr}
}
\label{tab:Quantitative_Table}
\end{table}
\squeezeup
\subsection{Ablation Study} 
By taking 512 x 384 resolution on the VITON-HD dataset as the basic setting, we conduct ablation studies to validate the effectiveness of several components in our networks, and the results are shown in Table \ref{tab:Tryon-condition}.

\begin{table}[b]
\caption{Condition effectiveness - ablation study on VITON-HD (512x384). We multiply KID by 100 for better comparison.}
\centering
\scalebox{0.8}{
\begin{tblr}{
  row{1} = {c},
  row{2} = {c},
  cell{1}{1} = {r=2}{},
  cell{1}{2} = {c=4}{},
  cell{3}{2} = {c},
  cell{3}{3} = {c},
  cell{3}{4} = {c},
  cell{3}{5} = {c},
  cell{4}{2} = {c},
  cell{4}{3} = {c},
  cell{4}{4} = {c},
  cell{4}{5} = {c},
  cell{5}{2} = {c},
  cell{5}{3} = {c},
  cell{5}{4} = {c},
  cell{5}{5} = {c},
  cell{6}{2} = {c},
  cell{6}{3} = {c},
  cell{6}{4} = {c},
  cell{6}{5} = {c},
  cell{7}{2} = {c},
  cell{7}{3} = {c},
  cell{7}{4} = {c},
  cell{7}{5} = {c},
  vline{2} = {1-7}{},
  hline{1,3,8} = {-}{},
}
\textbf{Method}                    & \textbf{VITON-HD – 512 x 384} &                 &               &                \\
                                   & LPIPS↓                       & SSIM↑           & FID↓          & KID↓           \\
w/o  Unconditional  Postprocessing & 0.0706                       & 0.9045          & 8.52          & 0.074          \\
w/o  Noise Condition Image         & 0.0923                       & 0.8774          & 11.00         & 0.21           \\
w/o  Global Condition              & 0.0759                       & 0.9036          & 9.11          & 0.11           \\
w/o Densepose Condition            & 0.1725                       & 0.8320          & 24.49         & 1.43           \\
Ours                               & \textbf{0.0675}              & \textbf{0.9091} & \textbf{8.43} & \textbf{0.066} 
\end{tblr}
}
\label{tab:Tryon-condition}
\end{table}
\begin{figure}[t]
\centering
\scalebox{0.8}{
\includegraphics[
    width=12cm,
  keepaspectratio,
]{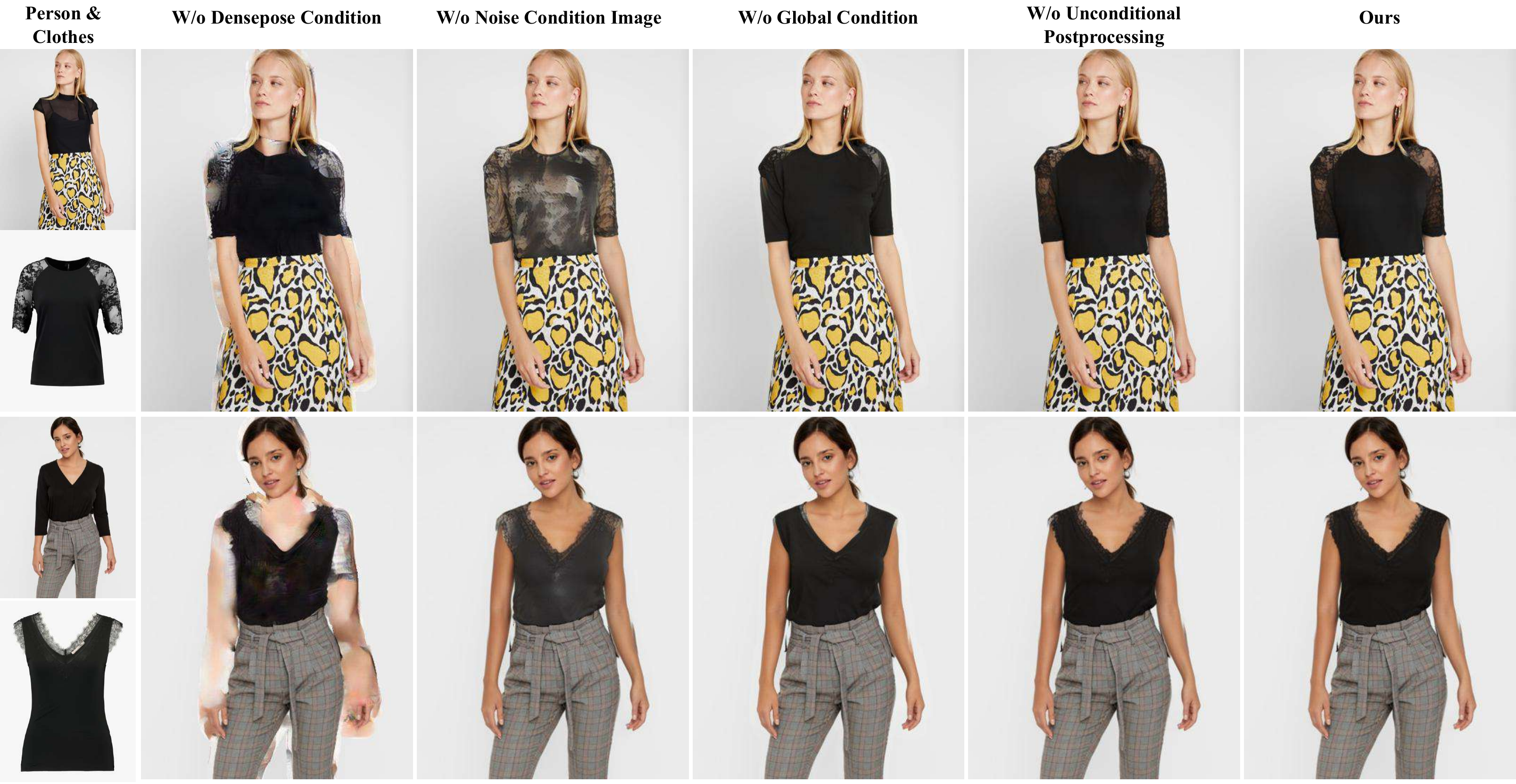}}
\caption{Visual condition effectiveness ablation studies for the Tryon module in our approach. Note that all ablation studies are applied to conditional post-processing. Please zoom in for better visualization.}
\label{fig: Cond-Ablation}
\end{figure}
\subsubsection{Condition Effectiveness - Tryon Module.}
In this ablation study, detailed in Table \ref{tab:Tryon-condition}, we systematically assess the components of our Tryon module. First, we explore the impact of removing the unconditional post-processing block. In this experiment, the local condition image \(I_c\) instead of \(I'_c\) is used to make the noise conditions image, we find it slightly degrades the model quality to imply that unconditional post-processing is still important. Second, we investigate the role of the generated noise condition (\(I'_n\)) by substituting it with regular Gaussian noise at the same level. Meanwhile, in the Global Condition effectiveness assessment, we just replaced the CLIP-embedding vector with the same shape zeros vector. In terms of dense pose conditions experiment, we do the same as the global condition case. Our findings underscore the significant influence of both the dense pose condition and the noise condition image, with the former being the most impactful, followed by the latter. Additionally, the global condition - the CLIP-based embedding module shows its importance as the third most influential block. Intriguingly, the removal of the first post-processing block in the Tryon module exhibits minimal impact on model quality. These results are visually demonstrated in Figure \ref{fig: Cond-Ablation}, where removal of the densepose condition (\emph{w/o Densepose Condition}) destroys body part structures and clothes, while the exclusion of the noise condition image results in pure noise (\emph{w/o Noise Condition Image}), undermining clothing texture. The model's failure to recognize and retain specific cloth textures, evident in the absence of the global condition (\emph{w/o Global Condition}) - the transparent sleeves or the frill of the cloth, highlights its significance. Interestingly, the decision not to apply unconditional post-processing results in only a slight reduction in output detail (\emph{w/o Unconditional Post-processing}).
\subsubsection{Additional Ablation Study.} Since the number of pages is limited, additional ablation experiment parts are shown in the Appendix. We answer the question \emph{"Does the advantage come from the modification of the Warping Module or come from the stronger prior?"}, \emph{"What is the trade-off between multiple and single timestep diffusion?"}, and \emph{"What is the impact of modification in the Warping Module?"}. Furthermore, there is also an ablation experiment to prove the ability of \emph{plug-and-play} mask-aware postprocessing block.

\squeezeup
\begin{figure}[t]
\centering
\scalebox{0.8}{\includegraphics[
    width=12cm,
  keepaspectratio,
]{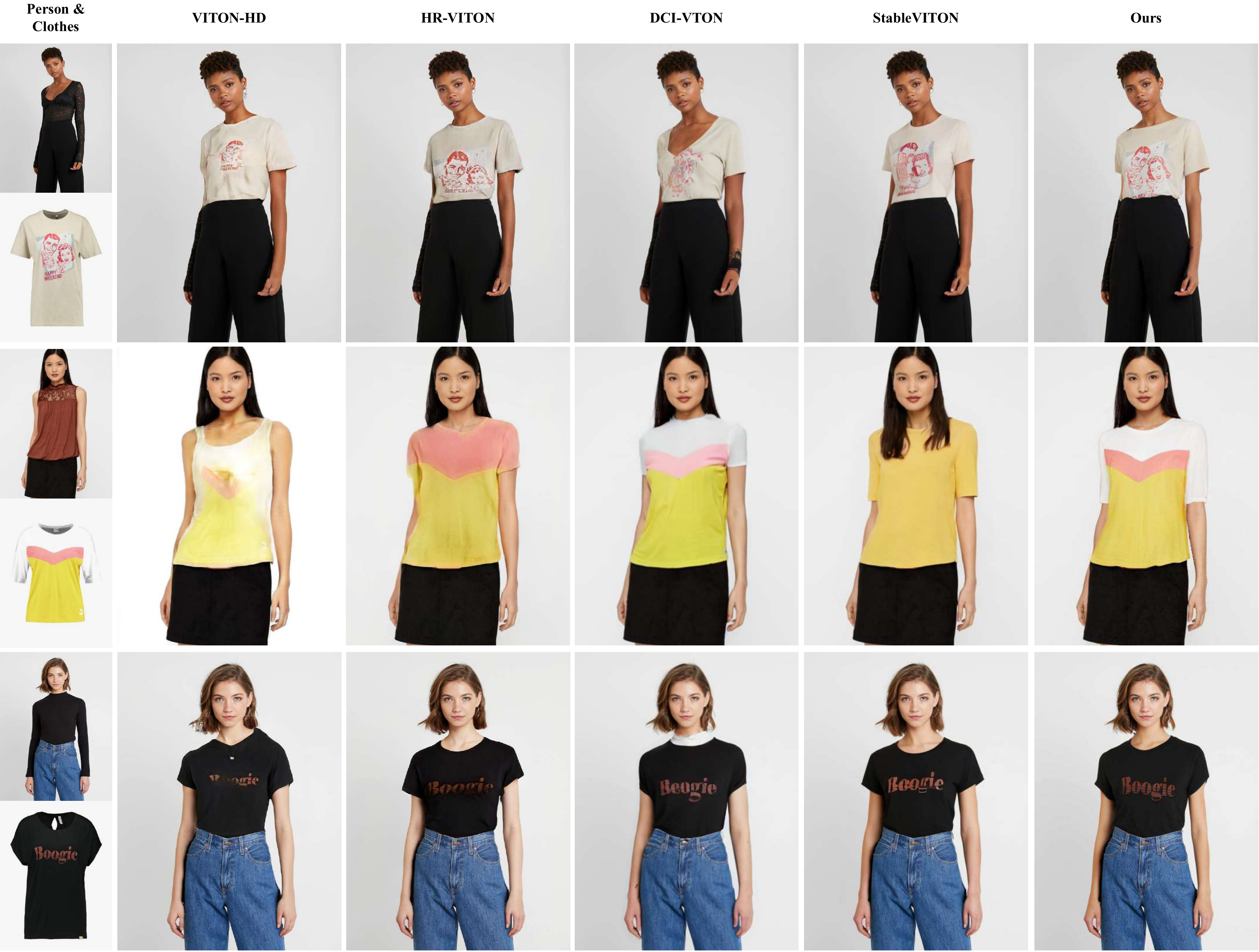}}
\centering
\caption{Qualitative comparison with baseline in VITON-HD Dataset \cite{VITON-HD} at 512 x 384 resolutions. Please zoom in for better visualization.}
\label{fig: Output-image-1}
\end{figure}

\subsection{Qualitative Evaluation}
In Fig. \ref{fig: Output-image-1}, we showcase composite images from various methods on the VITON-HD Dataset at 512 x 384 resolution. Our approach demonstrates superior performance to StableVITON, and DCI-VTON, the current SOTA results on the VITON-HD dataset regarding detail features. Our architecture generates highly realistic images, surpassing previous studies like VITON-HD and HR-VTON, particularly excelling in costume details compared to DCI-VTON and StableVITON. Notably, our model effectively handles challenging cases such as complicated symbols and characters on the clothes
the first and third examples, where DCI-VTON and StableVITON fall short. Moreover, in these cases, DCI-VTON introduces artifacts (neck part of the DCI-VTON result), while our approach maintains realism without these artifacts. Furthermore, the second row of Fig. \ref{fig: Output-image-1} showcases our outperforming in quality in both the scale and reality of the clothes compared with StableVITON and DCI-VTON. The additional qualitative results are shown in the Appendix.

\begin{figure}[t]
\centering
\scalebox{0.8}{\includegraphics[
    width=12cm,
  keepaspectratio,
]{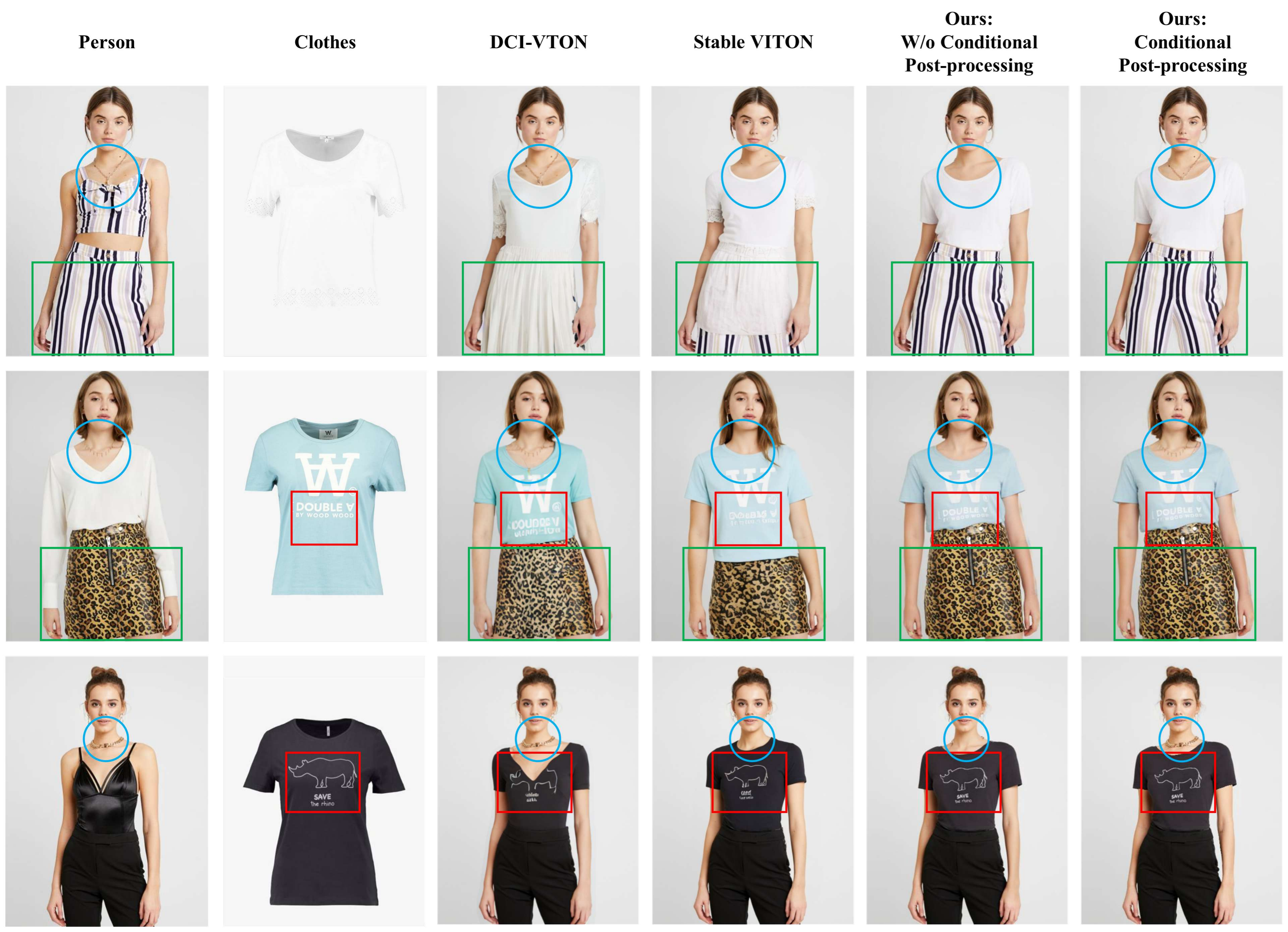}}
\caption{Example for the identity preservation on VTON-HD Dataset (512 x 384). The blue circle represents the important parts that need to be preserved, the green rectangle is for the wrong context compared to our approach, and the red rectangle is for the detailed part of the clothes that can not be retained compared to our approach. Please zoom in for better visualization.}
\label{fig: Identity-remain}
\end{figure}

Our research not only preserves outfit details and minimizes artifacts but also addresses the crucial issue of retaining identity information.  Leveraging our conditional mask-awareness post-processing technique, we successfully address this concern. Fig. \ref{fig: Identity-remain} illustrates the efficacy of our post-processing, comparing results before and after its application alongside the outcomes of the SOTA paper under the same conditions. Our technique effectively preserves identity information, demonstrated in cases where DCI-VTON fails to retain arm tattoos (Fig. \ref{fig: Output-15}). DCI-VTON introduces artifacts in the neck area (strange necklaces) and the lower-cloth context, compromising image authenticity (\(1^{st}\), \(2^{nd}\) rows ). Additionally, DCI-VTON and StableVITON struggle to maintain outfit details, \(2^{nd}\) and \(3^{rd}\) samples in Fig. \ref{fig: Identity-remain}. Our research ensures the maximum retention of identity information for the inference person without compromising output authenticity. The additional qualitative results are shown in the Appendix.
\squeezeup
\squeezeup
\section{Conclusions}
\squeezeup
In this study, we have addressed the challenges of Virtual Try-On technology, by introducing a novel diffusion-based approach that adeptly preserves garment texture and user identity. Our integrated system, comprising a warping module and a try-on module, enhanced with a mask-awareness post-processing technique, significantly outperforms existing methods in inference speed, being over 17.43 times faster than the current state-of-the-art, while maintaining superior fidelity in output. This dual focus on efficiency and detail preservation marks a substantial advancement in the field. However, it is important to acknowledge a limitation in our approach: the necessity for an elaborate post-processing process. While crucial for ensuring the integrity of the individual's identity and the garment's texture, this additional step adds complexity to the overall system. Despite this, the proposed method presents a promising solution for real-world applications, offering a more seamless and accurate virtual try-on experience. Future work could aim to streamline this post-processing phase, further enhancing the system's efficiency and applicability in diverse scenarios, including a wider range of clothing styles and body types.
\clearpage
\section*{Acknowledgments}
This research was supported by the MSIT(Ministry of Science and ICT), Korea, under the Grand Information Technology Research Center support program(IITP-2022-2020-0-01489) supervised by the IITP(Institute for Information \& communications Technology Planning \& Evaluation); and the Institute for Information \& Communications Technology
Promotion (IITP) grant funded by the Korean government (MSIP) (No. 2022-0-00407). 

\bibliographystyle{splncs04}
\bibliography{main}

\title{Time-Efficient and Identity-Consistent \\ Virtual Try-On Using A Variant of Altered Diffusion Models \\-- Appendix --}



\author{Phuong Dam \inst{1}\orcidlink{0009-0004-8422-3881} \and
Jihoon Jeong\inst{1}\orcidlink{0009-0003-0250-4296} \and
Anh Tran \inst{2}\orcidlink{0000-0002-3120-4036} \and 
Daeyoung Kim \inst{1}\orcidlink{0000-0002-7960-5955}}


\authorrunning{Phuong Dam, et al.}

\institute{Korea Advanced Institute of Science and Technology (KAIST) \\ \email{\{hoangphuong1211, jihooni, kimd\}@kaist.ac.kr} \and VinAI Research \\
\email{v.anhtt152@vinai.io}}

\maketitle
\renewcommand*{\thesection}{\Alph{section}}
\renewcommand*{\thesubsection}{\arabic{subsection}}
\renewcommand{\squeezeup}{\vspace{-2.0 mm}}

\section{Additional Ablation Study}
\subsection{Cross-Attention Effectiveness}
\begin{figure}[t]
\centering
\scalebox{0.8}{
\includegraphics[
    width=12cm,
  keepaspectratio,
]{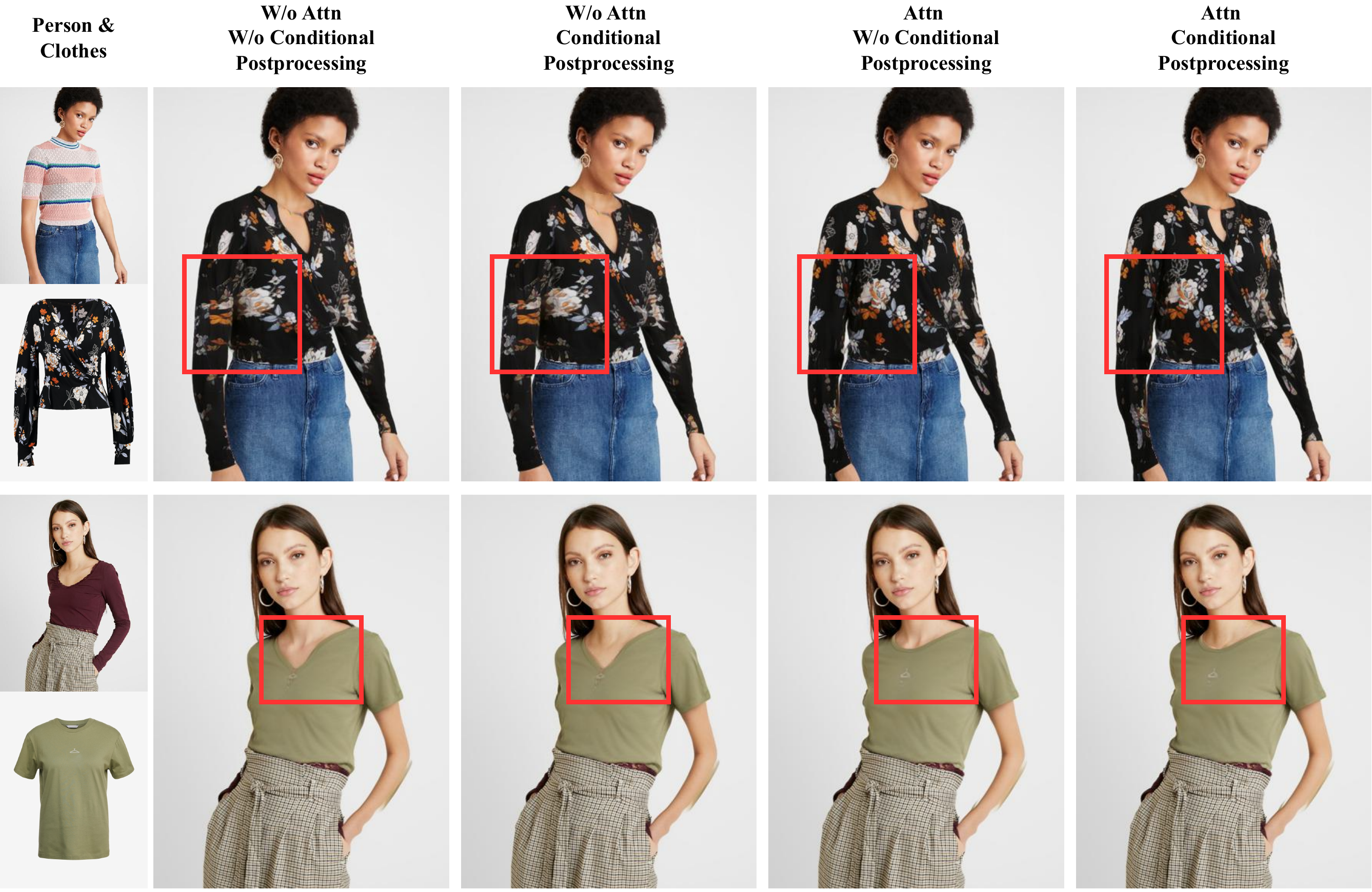}}
\caption{Visual cross-attention ablation studies in our approach. Please zoom in for better quality.}
\label{fig: Attn-Ablation}
\end{figure}

In this section, we will answer the question \emph{"What is the impact of modification in the Warping Module?"} by exploring how much the cross-attention mechanism affects the output of the model. Specifically, as mentioned above, we only apply the cross-attention in the stage of feature map resolution smaller than  64x48. In the case of removing the cross-attention block, we replace them with the concatenation operation. In this ablation study, we choose to compare the performance between the warping module with and without the multi-head cross-attention mechanism (\emph{W/o Attn}) under with and without the conditional post-processing block (\emph{W/o Conditional Post-processing}). Furthermore, we also measure the effectiveness of the post-processing block for both cases. Based on the result in Table \ref{tab:Cross attention_VTON-HD}, it can be seen that the cross-attention in the warping module provides a better result in almost all the comparison categories. However, applying cross-attention increases the number of parameters by 31\% (from 128.4M to 168.5M) with the resolution here being 512x384. We also visualized the qualitative improvement in Figure \ref{fig: Attn-Ablation}. It is observable that the absence of the cross-attention mechanism can result in distorted outputs from the warping module (as illustrated in the first example of Figure \ref{fig: Attn-Ablation}), potentially degrading the final visual result. Moreover, in the second example of Figure \ref{fig: Attn-Ablation}, the collar part of the garment is wrong-warped compared to the cross-attention ones, and the umbrella label also disappears in the non-cross-attention ones. This comparative analysis underscores the critical role of cross-attention in preserving detail and maintaining structural integrity in the warping process, thereby substantiating its inclusion despite the associated increase in computational complexity.
\squeezeup

\begin{table}[b]
\centering
\caption{Cross attention - ablation study on VITON-HD \cite{VITON-HD}(512x384). We multiply KID by 100 for better comparison. The number of parameters (nParam(M)) is excluded the CLIP vision embedding model}
\scalebox{0.85}{
\begin{tblr}{
  row{1} = {c},
  row{2} = {c},
  cell{1}{1} = {r=2}{},
  cell{1}{2} = {c=5}{},
  cell{3}{2} = {c},
  cell{3}{3} = {c},
  cell{3}{4} = {c},
  cell{3}{5} = {c},
  cell{3}{6} = {c},
  cell{4}{2} = {c},
  cell{4}{3} = {c},
  cell{4}{4} = {c},
  cell{4}{5} = {c},
  cell{4}{6} = {c},
  cell{5}{2} = {c},
  cell{5}{3} = {c},
  cell{5}{4} = {c},
  cell{5}{5} = {c},
  cell{5}{6} = {c},
  cell{6}{2} = {c},
  cell{6}{3} = {c},
  cell{6}{4} = {c},
  cell{6}{5} = {c},
  cell{6}{6} = {c},
  hline{1,3,7} = {-}{},
}
\textbf{Method}                 & \textbf{VITON-HD – (512 x 384)} &                 &                &                &                \\
                                & LPIPS↓                       & SSIM↑           & FID↓           & KID↓           & nParam(M)↓     \\
No\_Attn +
  w/o post-processing & 0.0691                       & 0.9015          & 8.55           & 0.076          & \textbf{128.4} \\
No\_Attn +
  post-processing     & \textbf{0.0670}              & 0.9074          & 8.52           & 0.074          & \textbf{128.4} \\
Attn
  + w/o post-processing     & 0.0706                       & 0.9016          & 8.49          & 0.071          & 168.5          \\
Attn
  + post-processing         & 0.0675                       & \textbf{0.9091} & \textbf{8.43} & \textbf{0.066} & 168.5          
\end{tblr}
}
\label{tab:Cross attention_VTON-HD}
\end{table}

\subsection{Noise Level Efficiency - Prior Assessment}
\begin{figure}[t]
\centering
\scalebox{0.8}{
\includegraphics[
    width=12cm,
  keepaspectratio,
]{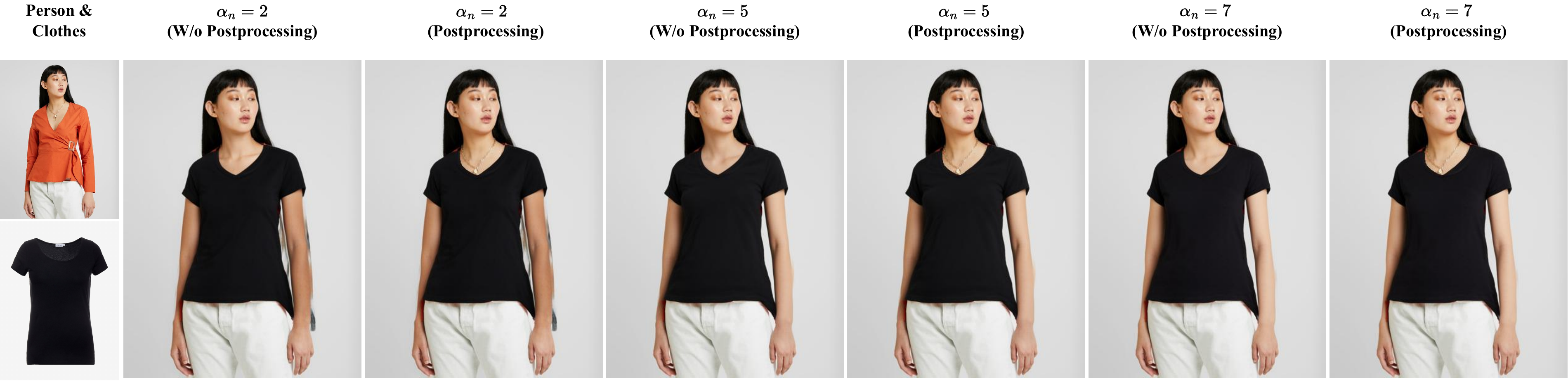}}
\caption{Visualize Noise level ablation studies. Please zoom in for better quality.}
\label{fig: Alpha-Ablation}
\end{figure}

This section explores the impact of varying noise levels on the performance of the Try-on Module. To ensure a controlled comparison, we use the same configuration setting for all \(\alpha_n\) values and the threshold of conditional mask-aware post-processing will still be 0.8 in this ablation study. The noise levels tested in this study are represented by \(\alpha_n\) values of 2, 5, and 7. As depicted in Table \ref{tab:Noise Ablation Study}, our Try-on Module works best under \(\alpha_n = 5\). In addition, if we add too little noise (low level of noise - \(\alpha_n = 2\)), the try-on module performance will decrease massively. Meanwhile, when the level of noise is too high \(\alpha_n = 7\), the Try-on model performance also be affected - slightly reduced. As depicted in Fig. \ref{fig: Alpha-Ablation}, it is easy to see that when the \(\alpha_n = 2\) the skin color output looks not real and several artifacts also appear in this experiment. However, when we increase the value of \(\alpha_n\) to 5 or 7, the generated image looks more realistic.    

Combining with the above assessment, we can answer the question \emph{"Does the advantage come from the modification of the Warping Module or come from the stronger prior?"} that the proof of our positive impact comes from \emph{stronger prior}.     
\squeezeup

\begin{table}[b]
\centering
\caption{Noise Level ablation study on VITON-HD \cite{VITON-HD}(512x384). We multiply KID by 100 for better comparison}
\label{tab:Noise Ablation Study}
\scalebox{0.8}{
\begin{tblr}{
  row{1} = {c},
  row{2} = {c},
  cell{1}{1} = {r=2}{},
  cell{1}{2} = {c=4}{},
  cell{3}{2} = {c},
  cell{3}{3} = {c},
  cell{3}{4} = {c},
  cell{3}{5} = {c},
  cell{4}{2} = {c},
  cell{4}{3} = {c},
  cell{4}{4} = {c},
  cell{4}{5} = {c},
  cell{5}{2} = {c},
  cell{5}{3} = {c},
  cell{5}{4} = {c},
  cell{5}{5} = {c},
  cell{6}{2} = {c},
  cell{6}{3} = {c},
  cell{6}{4} = {c},
  cell{6}{5} = {c},
  cell{7}{2} = {c},
  cell{7}{3} = {c},
  cell{7}{4} = {c},
  cell{7}{5} = {c},
  cell{8}{2} = {c},
  cell{8}{3} = {c},
  cell{8}{4} = {c},
  cell{8}{5} = {c},
  hline{1,3,9} = {-}{},
  hline{2} = {2-5}{},
}
\textbf{Method}                  & \textbf{VITON-HD - (512x384)} &                 &               &                \\
                                 & LPIPS↓                     & SSIM↑           & FID↓          & KID↓           \\
\(\alpha_n = 2\) (w/o post-processing) & 0.0924                     & 0.8854          & 9.54          & 0.173          \\
\(\alpha_n = 2\) (post-processing)     & 0.0915                     & 0.8903          & 9.61          & 0.175          \\
\(\alpha_n = 5\) (w/o post-processing) & 0.0706                     & 0.9016          & 8.49          & 0.071          \\
\(\alpha_n = 5\) (post-processing)     & \textbf{0.0675}            & \textbf{0.9091} & \textbf{8.43} & \textbf{0.066} \\
\(\alpha_n = 7\) (w/o post-processing) & 0.0712                     & 0.9017          & 8.77          & 0.095          \\
\(\alpha_n = 7\) (post-processing)     & 0.0681                     & \textbf{0.9091} & 8.74          & 0.094          
\end{tblr}}
\end{table}
\subsection{Multiple and Single Time-Step Diffusion Trade-Off}
Choosing the Single-Time Step as the main approach for the Try-on Module, there is a concern about whether it reduces the quality of the images instead. To answer the question \emph{"What is the trade-off between multiple and single timestep diffusion?"}, we compared our single-step approach and a multiple-step diffusion approach. In this section, we trained our Tryon Module as a vanilla diffusion model with DDIM schedulers 1000 timesteps (DDIM*), and CLIP embedding is added to time embedding, the results of this experiment are shown in Table \ref{tab:Multiple and Single Time-Step Diffusion Trade-Off}. Our approach has a big gap in all categories. The reason may come from the difference in the objective function. While the DDIM* approach can only apply L2 loss in their noise prediction, ours can use L1, perceptual loss VGG \cite{PerceptualLoss}, and even adversarial loss (GAN Loss) \cite{RelativisticGAN}. Furthermore, the gap in the results also may come from the difference in the time spent for hyper-parameter tuning and the difference in prior.

\begin{table}[b]
\centering
\caption{Multiple and Single Time-Step Diffusion Trade-Off - ablation study on VITON-HD \cite{VITON-HD}(512x384). DDIM* is the vanilla diffusion model with DDIM schedulers 1000 timesteps. We multiply KID by 100 for better comparison.}
\scalebox{0.85}{
\begin{tblr}{
  row{1} = {c},
  row{2} = {c},
  cell{1}{1} = {r=2}{},
  cell{1}{2} = {c=5}{},
  cell{3}{2} = {c},
  cell{3}{3} = {c},
  cell{3}{4} = {c},
  cell{3}{5} = {c},
  cell{3}{6} = {c},
  cell{4}{2} = {c},
  cell{4}{3} = {c},
  cell{4}{4} = {c},
  cell{4}{5} = {c},
  cell{4}{6} = {c},
  cell{5}{2} = {c},
  cell{5}{3} = {c},
  cell{5}{4} = {c},
  cell{5}{5} = {c},
  cell{5}{6} = {c},
  cell{6}{2} = {c},
  cell{6}{3} = {c},
  cell{6}{4} = {c},
  cell{6}{5} = {c},
  cell{6}{6} = {c},
  hline{1,3,7} = {-}{},
}
\textbf{Method}                 & \textbf{VITON-HD – (512 x 384)} &                 &                &                &                \\
                                & LPIPS↓                       & SSIM↑           & FID↓           & KID↓           & T(s)↓     \\
DDIM* +
  w/o post-processing & 0.247              & 0.7978          & 14.05           & 0.468          & 460 \\
DDIM* +
  post-processing     & 0.241              & 0.8053          & 13.93           & 0.441          & 460 \\
Ours
  + w/o post-processing     & 0.0706                       & 0.9016          & 8.49          & 0.071          & \textbf{1.01}          \\
Ours
  + post-processing         & \textbf{0.0675}                       & \textbf{0.9091} & \textbf{8.43} & \textbf{0.066} & \textbf{1.01}          
\end{tblr}
}
\label{tab:Multiple and Single Time-Step Diffusion Trade-Off}
\end{table}

\squeezeup
\subsection{Conditional Mask-Aware Threshold}
\begin{figure}[t]
\centering
\scalebox{0.8}{
\includegraphics[
    width=12cm,
  keepaspectratio,
]{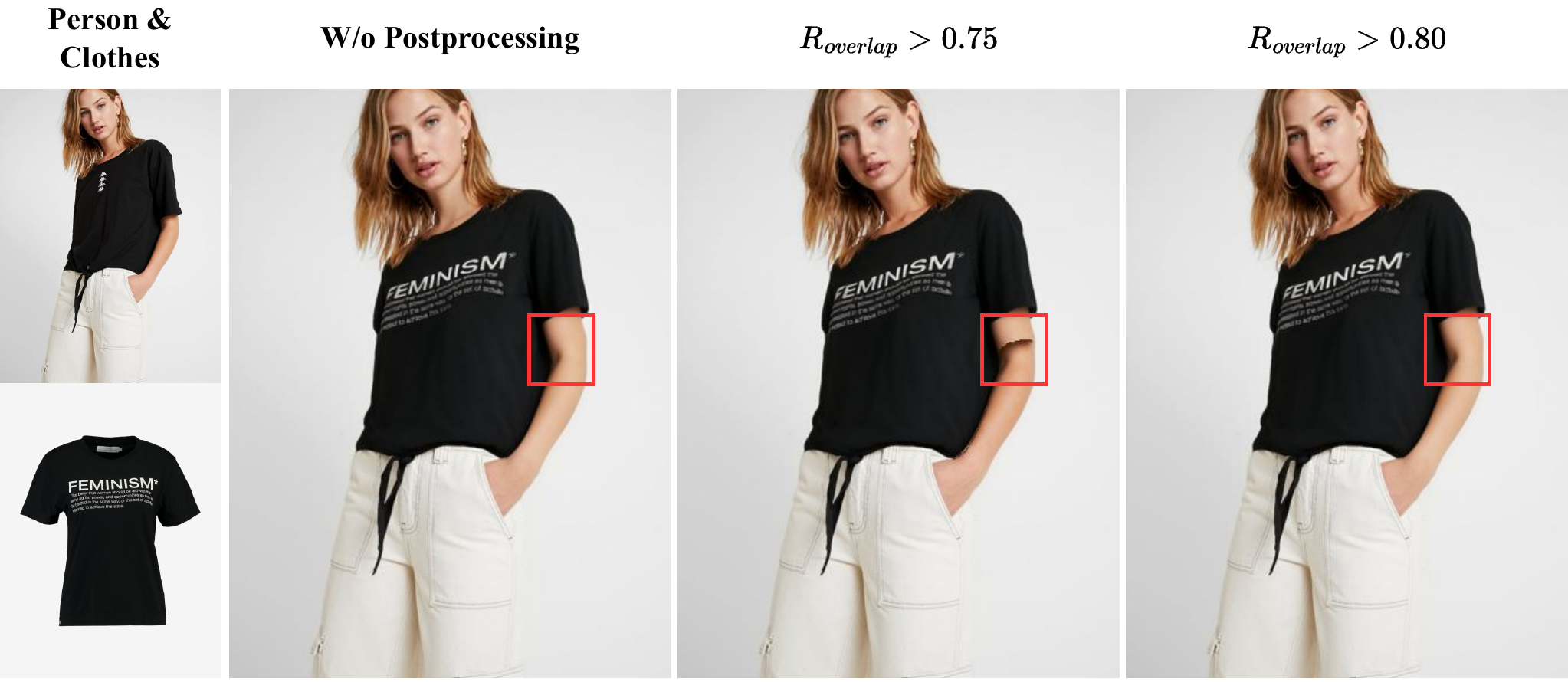}}
\caption{Visualize Conditional Mask-Aware Threshold ablation studies in our approach. Please zoom in for better quality.}
\label{fig: Overlap-Ratio-Ablation}
\end{figure}

In this analysis, we explore the effectiveness of the threshold value for conditional mask-aware post-processing. It is important to note that the experimental setup remains consistent with the standard implementation, with the only modification being the adjustment of the threshold for the overlapping ratio condition applied during post-processing.   Although the ideal theoretical value for this ratio in this algorithm is 1, empirical evidence presented in Table \ref{tab:Overlap Ratio} indicates that the most favorable quantitative outcomes are achieved when \(R_{overlap} > 0.75\), with the second-best performance observed at a threshold ratio of 0.8. However, because the value of 0.75 is too sensitive to the imperfection of the segmentation prediction, we chose the marginally less optimal but more robust threshold of 0.8 for our investigation. Specifically, it is obvious in Fig. \ref{fig: Overlap-Ratio-Ablation} that there is an artifact in the arm part (highlighted in red rectangle region) when the threshold value is just 0.75, and there is not if it is 0.8. This observation underscores the delicate balance between achieving optimal overlap ratios and mitigating the risk of artifacts due to segmentation imperfections.
\squeezeup

\begin{table}[t]
\centering
\caption{Overlap ratio \(R_{overlap}\) condition for mask-aware post-processing ablation study on VITON-HD \cite{VITON-HD}(512x384). The applying rate AR(\%) is represented for the rate of post-processing applied in at least one part of predicted body segmentation in the test set under unpaired configuration. We multiply KID by 100 for better comparison. \textbf{Bold} and \underline{underline} denote the best and the second best}
\label{tab:Overlap Ratio}
\scalebox{0.8}{
\begin{tblr}{
  row{1} = {c},
  row{2} = {c},
  cell{1}{1} = {r=2}{},
  cell{1}{2} = {c=5}{},
  cell{3}{2} = {c},
  cell{3}{3} = {c},
  cell{3}{4} = {c},
  cell{3}{5} = {c},
  cell{3}{6} = {c},
  cell{4}{2} = {c},
  cell{4}{3} = {c},
  cell{4}{4} = {c},
  cell{4}{5} = {c},
  cell{4}{6} = {c},
  cell{5}{2} = {c},
  cell{5}{3} = {c},
  cell{5}{4} = {c},
  cell{5}{5} = {c},
  cell{5}{6} = {c},
  cell{6}{2} = {c},
  cell{6}{3} = {c},
  cell{6}{4} = {c},
  cell{6}{5} = {c},
  cell{6}{6} = {c},
  cell{7}{2} = {c},
  cell{7}{3} = {c},
  cell{7}{4} = {c},
  cell{7}{5} = {c},
  cell{7}{6} = {c},
  cell{8}{2} = {c},
  cell{8}{3} = {c},
  cell{8}{4} = {c},
  cell{8}{5} = {c},
  cell{8}{6} = {c},
  cell{9}{2} = {c},
  cell{9}{3} = {c},
  cell{9}{4} = {c},
  cell{9}{5} = {c},
  cell{9}{6} = {c},
  hline{1,3,10} = {-}{},
  hline{2} = {2-6}{},
}
\textbf{Method}     & \textbf{VITON-HD – (512 x 384)} &                 &                &                &                 \\
                    & LPIPS↓                          & SSIM↑           & FID↓           & KID↓           & AR(\%)         \\
w/o post-processing  & 0.0706                          & 0.9016          & 8.492          & 0.071          & 0               \\
$R_{overlap} > 0.75$ & \textbf{0.0673}                 & \textbf{0.9094} & \textbf{8.423} & \underline{0.067}  & 83.19       \\
$R_{overlap} > 0.80$ & \underline{0.0675}             & \underline{0.9091}  & \underline{8.428}  & \textbf{0.066} & 79.63    \\
$R_{overlap} > 0.85$ & 0.0677                          & 0.9086          & 8.434          & \underline{0.067}  & 76.07       \\
$R_{overlap} > 0.90$ & 0.0680                          & 0.9076          & 8.452          & \textbf{0.066} & 70.77           \\
$R_{overlap} > 0.95$ & 0.0690                          & 0.9053          & 8.453          & 0.068          & 60.14           \\
$R_{overlap} = 1$ & 0.0706                          & 0.9017          & 8.464          & 0.069          & 17.16           
\end{tblr}}
\end{table}
\squeezeup

\subsection{Conditional Mask-aware Post-processing Performance.}
In addition, we not only evaluate the effectiveness of our "plug-and-play" post-processing block for our study but also do on previous studies that have the same network structure to prove its efficiency. Table \ref{tab:Post processing_previous} illustrates that our mask-aware postprocessing block can enhance the model performance in not only our study but also all previous studies under any constraint of resolution. Note that, the result in this table is experienced based on the pre-trained model provided by the research author, which is run on the same setting as in our inference time calculation. 

\begin{table}[b]
\caption{Conditional Post-processing on ours and other previous studies. The * represents that we only ran the experiment based on the pre-trained model at 1024 x 768 resolution published by the authors of these studies, then plugged in our post-processing technique. Our approach was applied to images of 512x384 resolution, consistent with the resolution used in the referenced experiments. We multiply KID by 100 for better comparison.}
\centering
\scalebox{0.8}{
\begin{tblr}{
  row{1} = {c},
  row{2} = {c},
  cell{1}{1} = {r=2}{},
  cell{1}{2} = {c=4}{},
  cell{3}{2} = {c},
  cell{3}{3} = {c},
  cell{3}{4} = {c},
  cell{3}{5} = {c},
  cell{4}{2} = {c},
  cell{4}{3} = {c},
  cell{4}{4} = {c},
  cell{4}{5} = {c},
  cell{5}{2} = {c},
  cell{5}{3} = {c},
  cell{5}{4} = {c},
  cell{5}{5} = {c},
  cell{6}{2} = {c},
  cell{6}{3} = {c},
  cell{6}{4} = {c},
  cell{6}{5} = {c},
  cell{7}{2} = {c},
  cell{7}{3} = {c},
  cell{7}{4} = {c},
  cell{7}{5} = {c},
  cell{8}{2} = {c},
  cell{8}{3} = {c},
  cell{8}{4} = {c},
  cell{8}{5} = {c},
  vline{2} = {-}{},
  hline{1,3,5,7,9} = {-}{},
}
\textbf{Method}                    & \textbf{VITON-HD }                             &                             &                         &                          \\
                                   & LPIPS↓                                         & SSIM↑                       & FID↓                    & KID↓                     \\
VITON-HD*  + w/o postprocessing    & 0.1308                                         & 0.8665                      & 11.755                  & 0.284                    \\
VITON-HD*  + postprocessing        & \textbf{0.1296}                                & \textbf{0.8691}             & \textbf{11.754}         & \textbf{0.282}           \\
HR-VTON*  + w/o postprocessing     & 0.1106                                         & 0.8816                      & 11.204                  & 0.268                    \\
HR-VTON*  + postprocessing         & \textbf{0.1068}                                & \textbf{0.8877}             & \textbf{11.195}         & \textbf{0.264}           \\
Ours + w/o postprocessing          & 0.0706                                         & 0.9016                      & 8.49                    & 0.071                    \\
Ours + postprocessing              &  \textbf{\textbf{0.0675}}                      & \textbf{\textbf{0.9091}}    & \textbf{\textbf{8.43}}  & \textbf{\textbf{0.066}}
\end{tblr}
}
\label{tab:Post processing_previous}
\end{table}
\squeezeup
\section{Additional Qualitative Results}
\subsection{Results on VITON-HD.} 
\begin{figure}
\centering
\includegraphics[
    width=12cm,
  keepaspectratio,
]{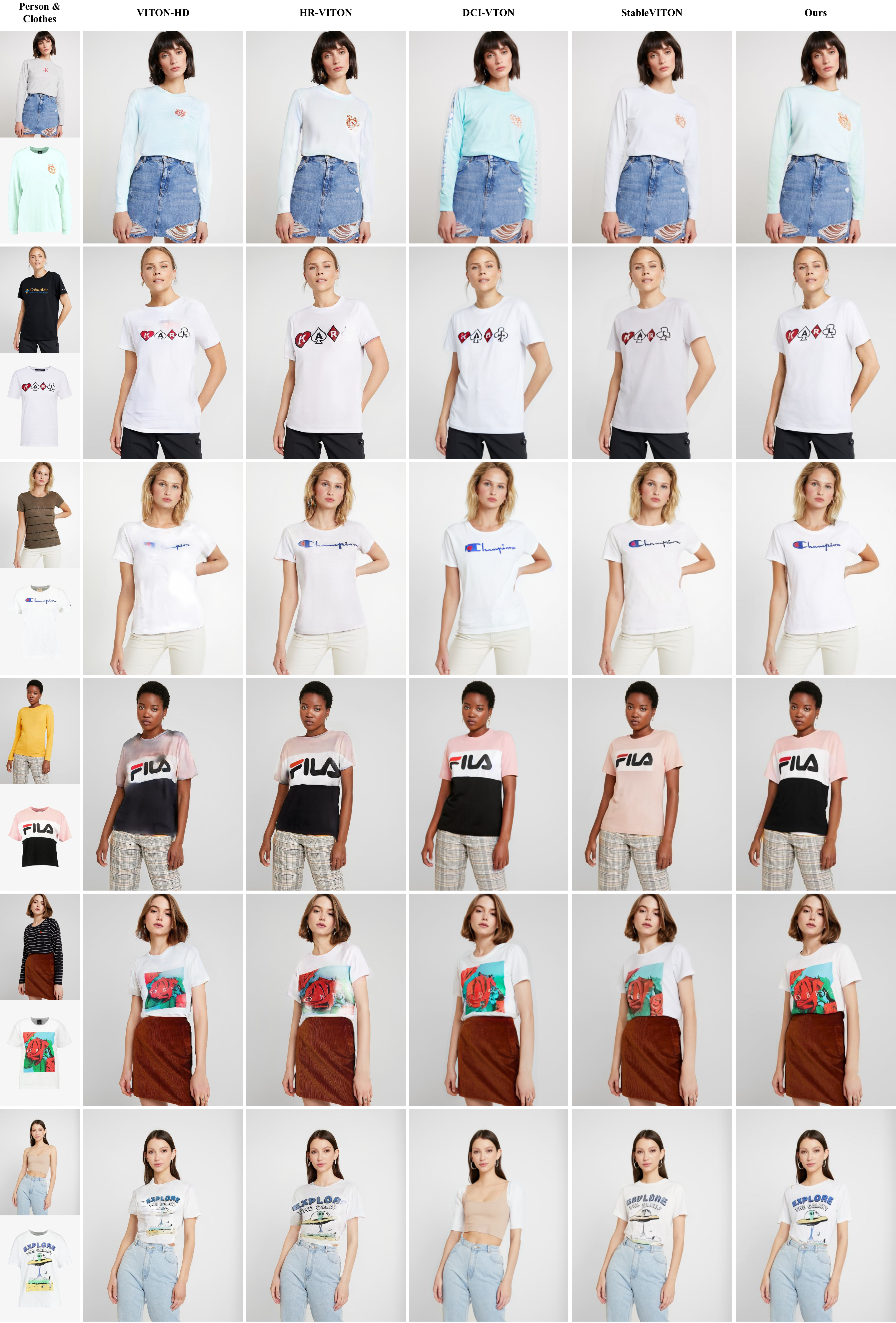}
\caption{Qualitative comparison of difference methods on VITON-HD Dataset. Please zoom in for better visualization.}
\label{fig: Output-image-2}
\end{figure}

\begin{figure}
\centering
\includegraphics[
    width=12cm,
  keepaspectratio,
]{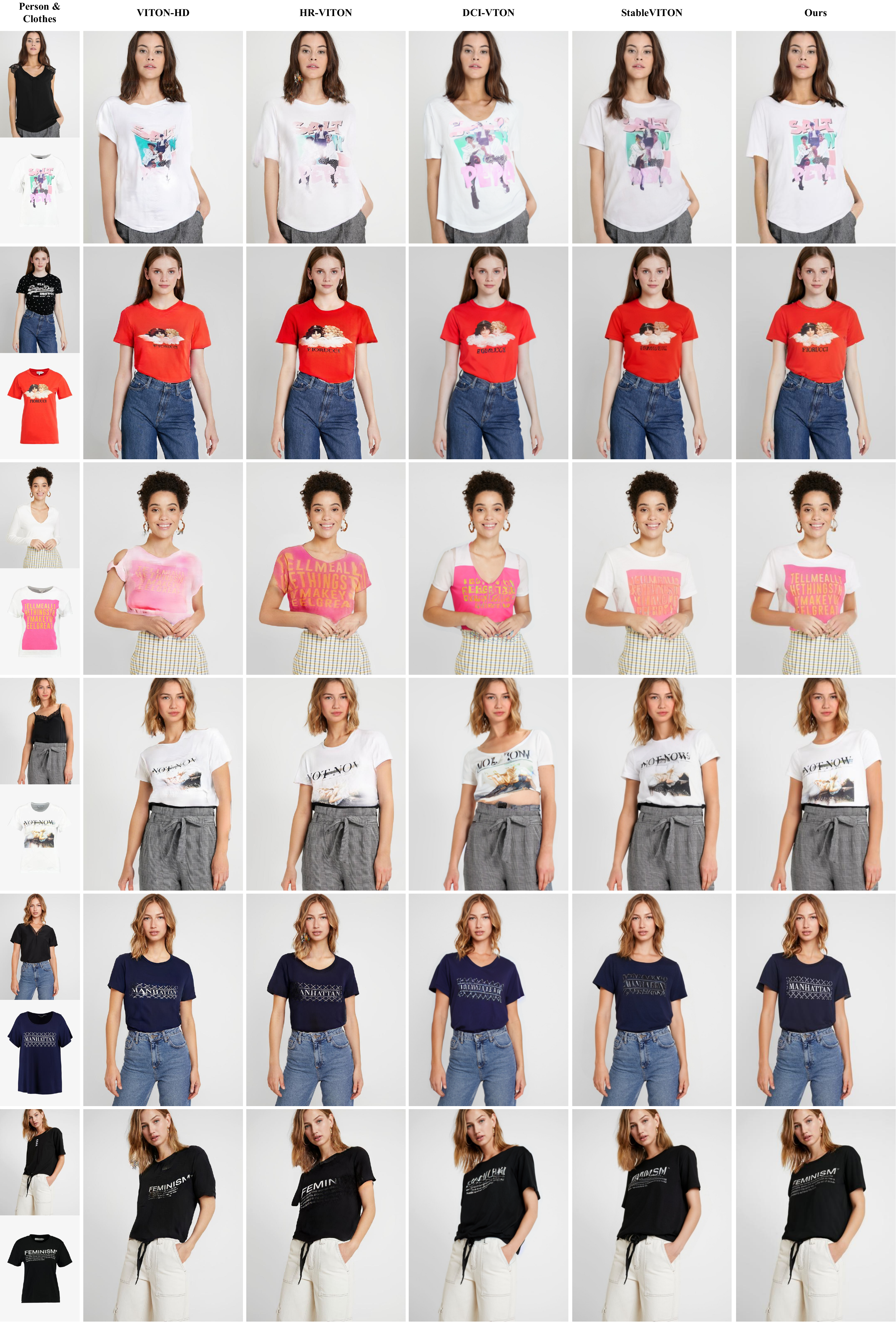}
\caption{Qualitative comparison of difference methods on VITON-HD Dataset. Please zoom in for better visualization.}
\label{fig: Output-image-3}
\end{figure}

This section presents additional composite images generated by various techniques on the VITON-HD dataset, showcased in Fig. \ref{fig: Output-image-2} and Fig. \ref{fig: Output-image-3}. The methods compared include VITON-HD \cite{VITON-HD}, HR-VTON \cite{HR-VTON}, DCI-VTON \cite{DCI-VTON}, and StableVITON \cite{Stableviton}. Our approach surpasses these prior methods, particularly excelling in synthesizing images that faithfully preserve clothing details, notably outperforming DCI-VTON and StableVITON.
Evidently, in the first, third, and fourth rows of Figure \ref{fig: Output-image-3}, DCI-VTON inaccurately depicts the context of the clothes collar, a shortcoming effectively addressed by our model. Moreover, instances last row of Figure \ref{fig: Output-image-2} reveal alterations in clothing style by DCI-VTON, whereas our model consistently maintains output fidelity. 
DCI-VTON and StableVITON also fail to ensure the detailed features of the garment (colors, symbols, and figures) (as seen in Figure \ref{fig: Output-image-2} and Figure \ref{fig: Output-image-3}), a challenge effectively addressed by our model. 

\begin{figure}
\centering
\includegraphics[
    width=12cm,
  keepaspectratio,
]{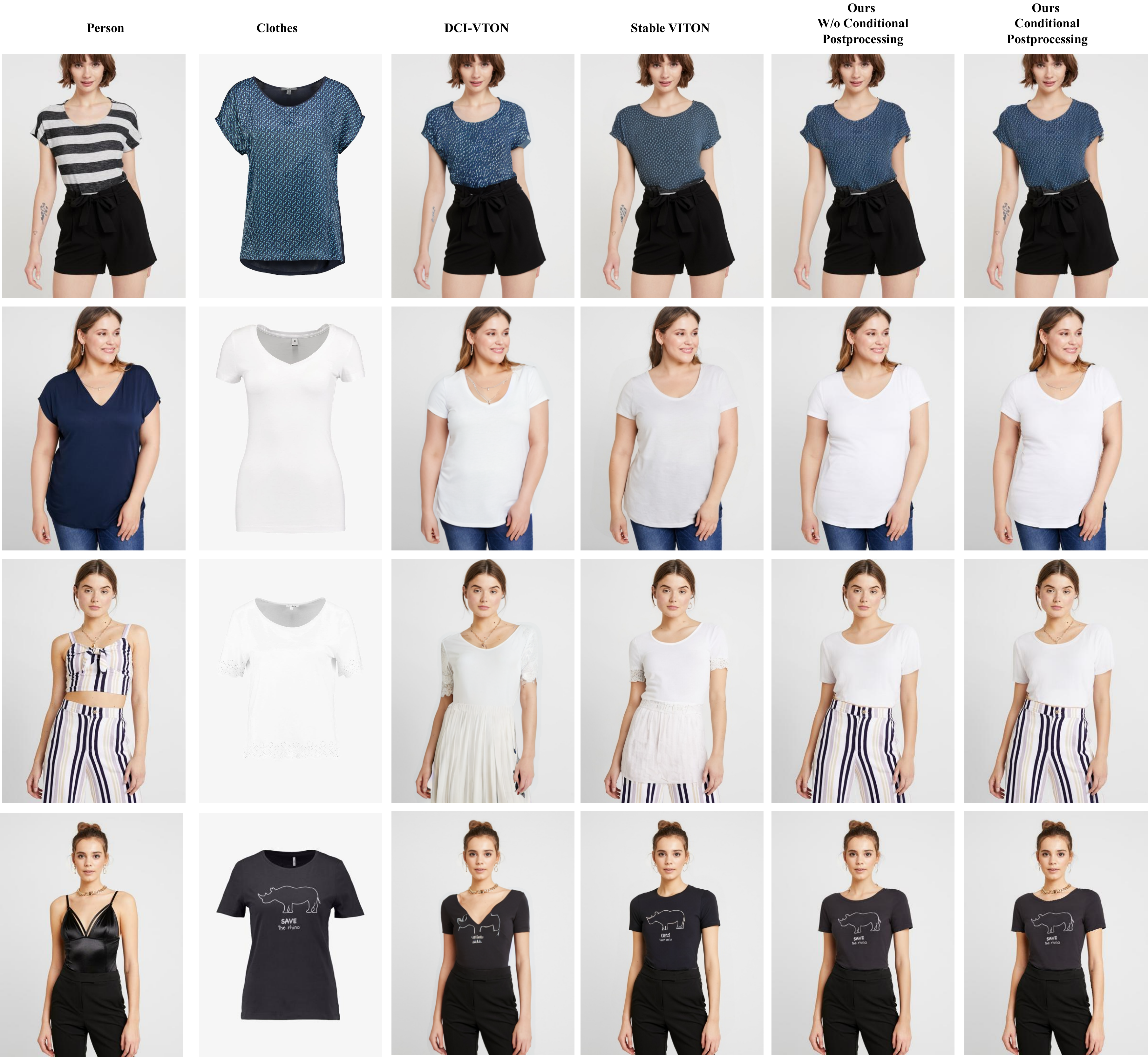}
\caption{Examples of unmarked version for the identity preservation on VITON-HD.}
\label{fig: Identity-2}
\end{figure}

\begin{figure}
\centering
\includegraphics[
    width=12cm,
  keepaspectratio,
]{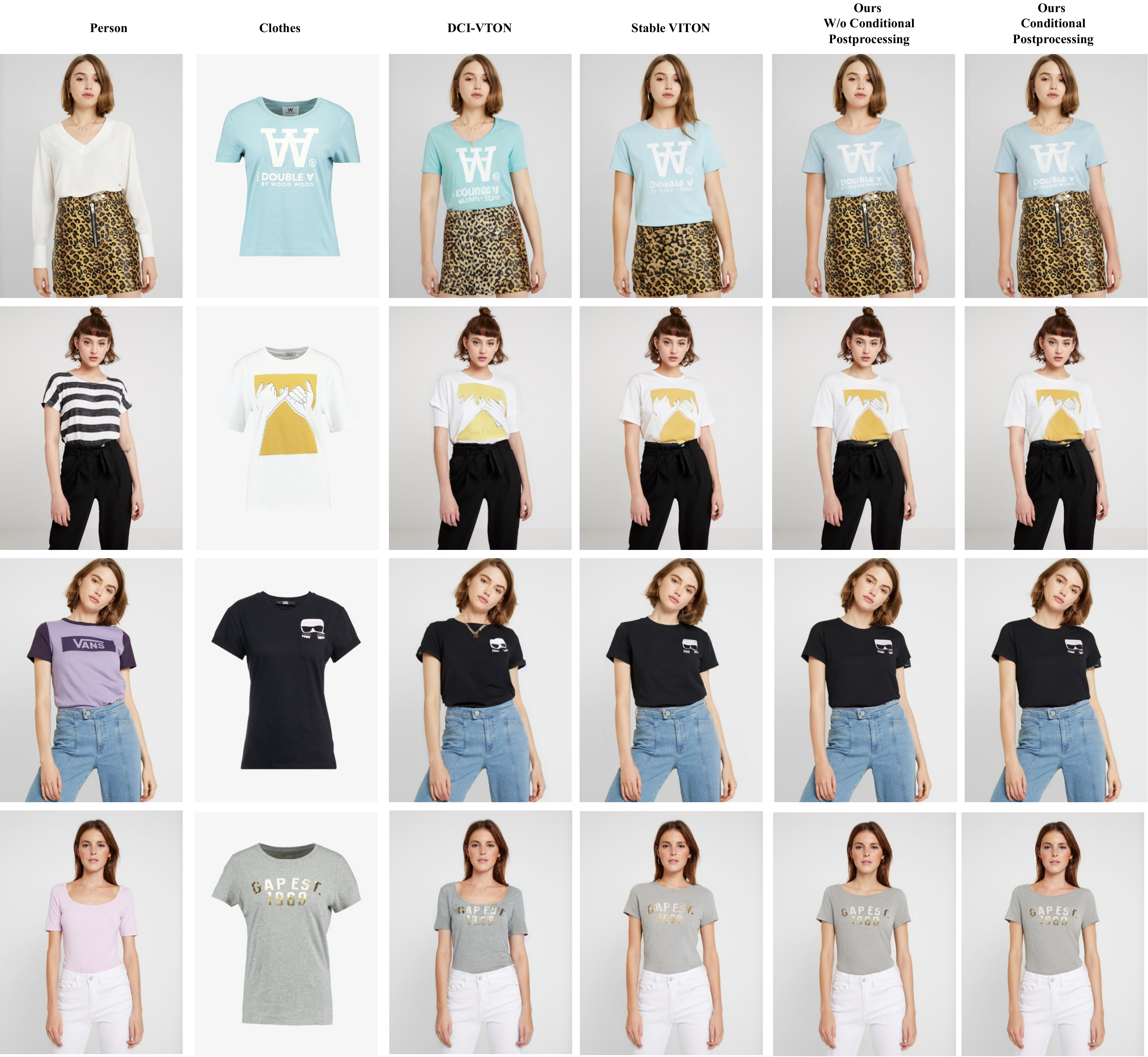}
\caption{Examples of unmarked version for the identity preservation on VITON-HD.}
\label{fig: Identity-3}
\end{figure}

Additionally, for a comprehensive understanding of the identity preservation aspect, the unmarked version of identities remaining and additional examples are visualized in Figure \ref{fig: Identity-2} and Figure \ref{fig: Identity-3}.
\subsection{Results on DressCode.}  
\begin{figure}
\centering
\includegraphics[
    width=12cm,
  keepaspectratio,
]{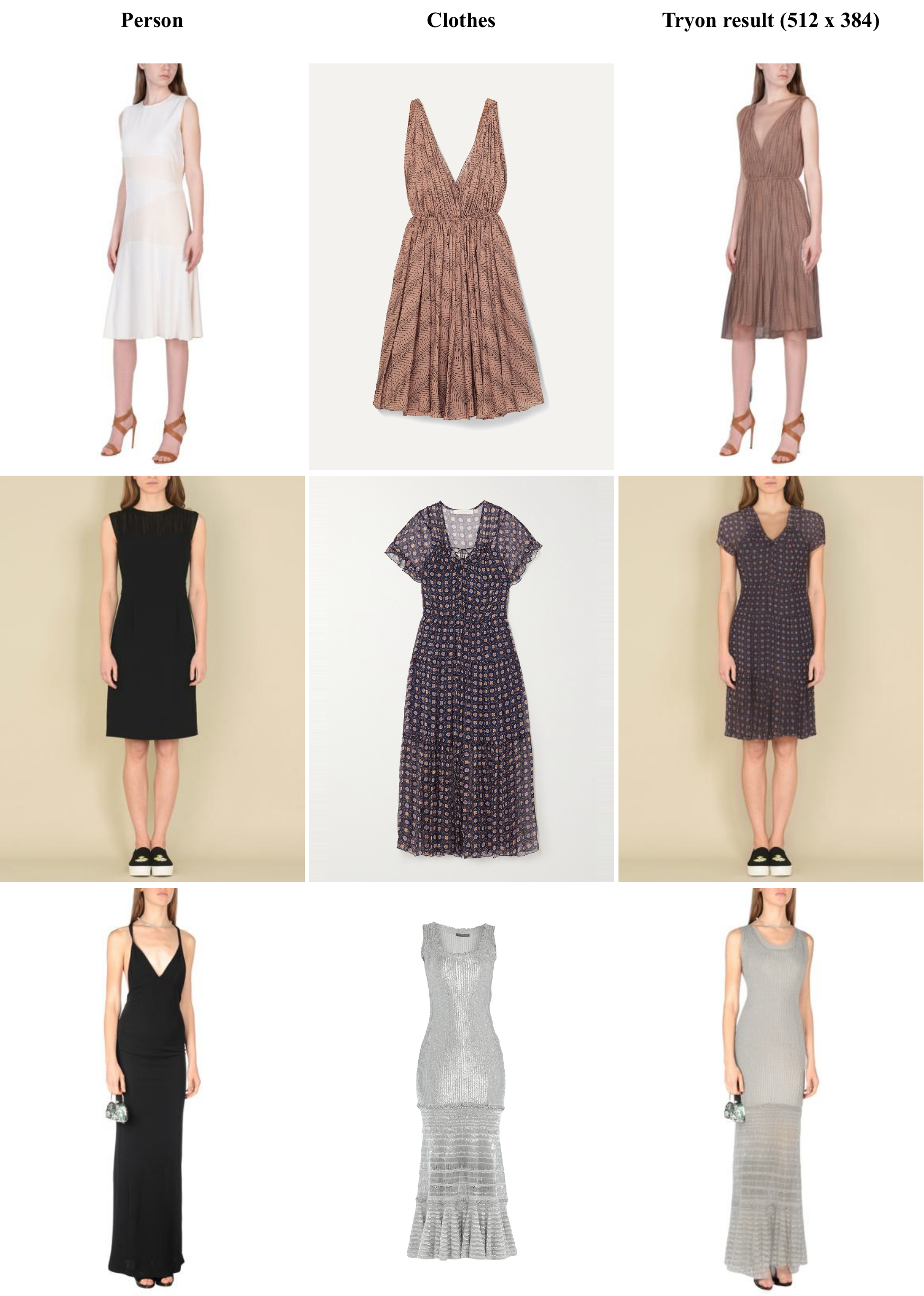}
\caption{Visualization results on DressCode Dataset (512 x 384) - Dress.}
\label{fig: Output-image-DressCode-1}
\end{figure}

\begin{figure}
\centering
\includegraphics[
    width=12cm,
  keepaspectratio,
]{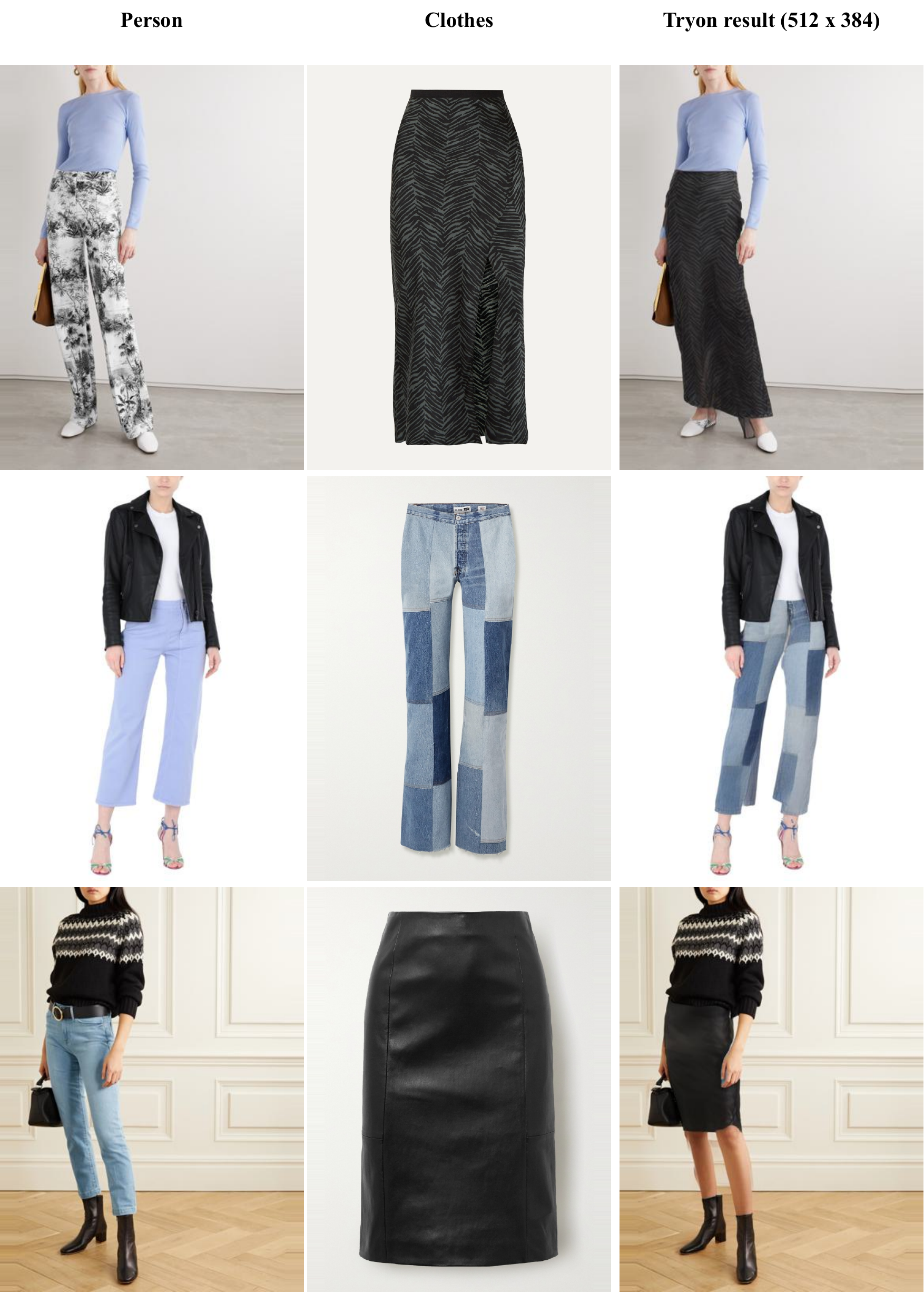}
\caption{Visualization results on DressCode Dataset (512 x 384) - Lower Clothes.}
\label{fig: Output-image-DressCode-2}
\end{figure}

\begin{figure}
\centering
\includegraphics[
    width=12cm,
  keepaspectratio,
]{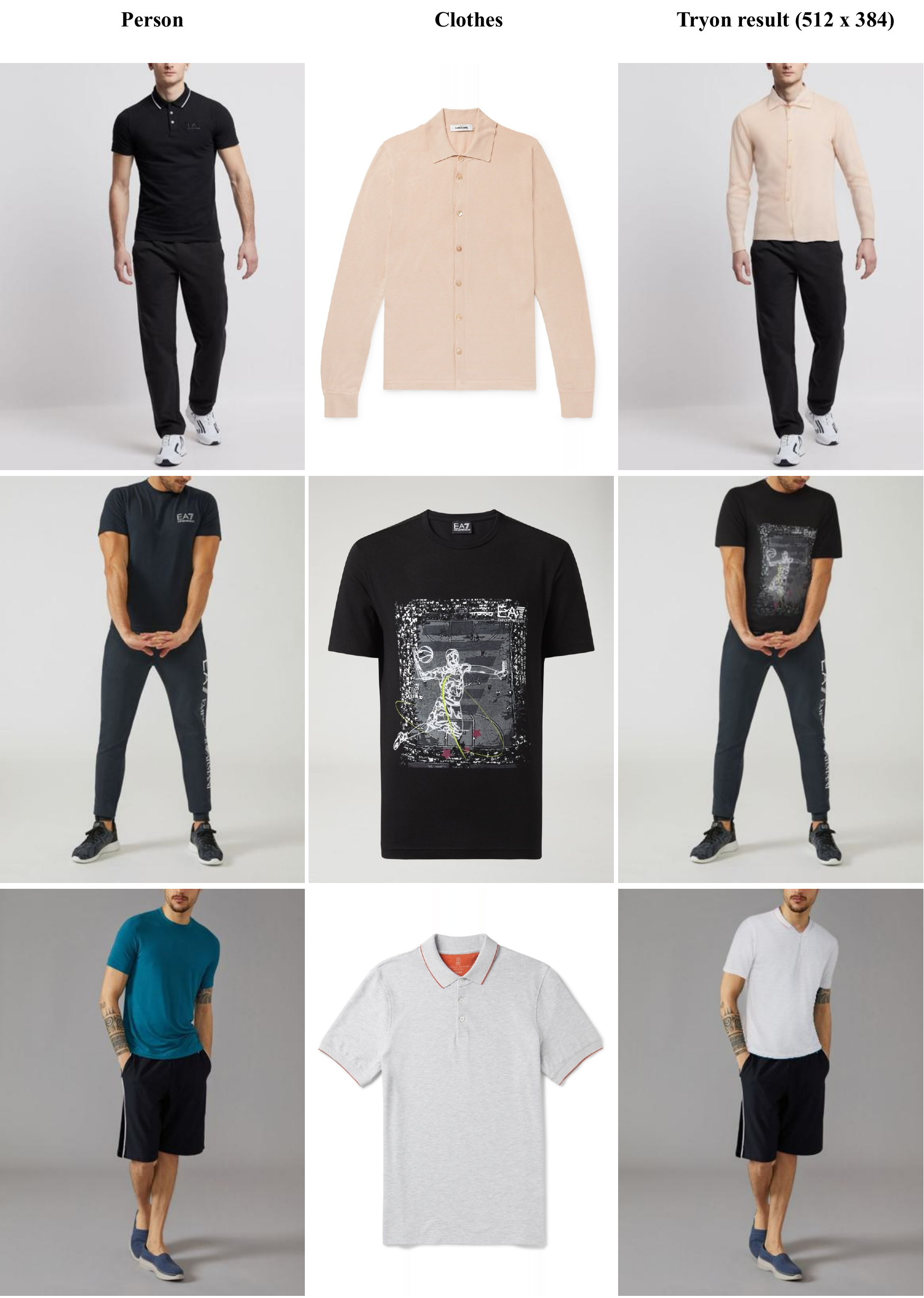}
\caption{Visualization results on DressCode Dataset (512 x 384) - Upper Clothes.}
\label{fig: Output-image-DressCode-3}
\end{figure}

In addition to quantitative assessments, we visually showcase the outcomes of our method in Figure \ref{fig: Output-image-DressCode-1}, Figure \ref{fig: Output-image-DressCode-2}, and Figure \ref{fig: Output-image-DressCode-3}. Across these sub-datasets, our method consistently achieves realistic and natural try-on results, effectively preserving unique identity details of the subject, such as muscle structure, tattoos, and necklaces.

\begin{figure}[b]
\centering
\scalebox{0.8}{\includegraphics[
  width=12cm,
  height=6cm,
  keepaspectratio,
]{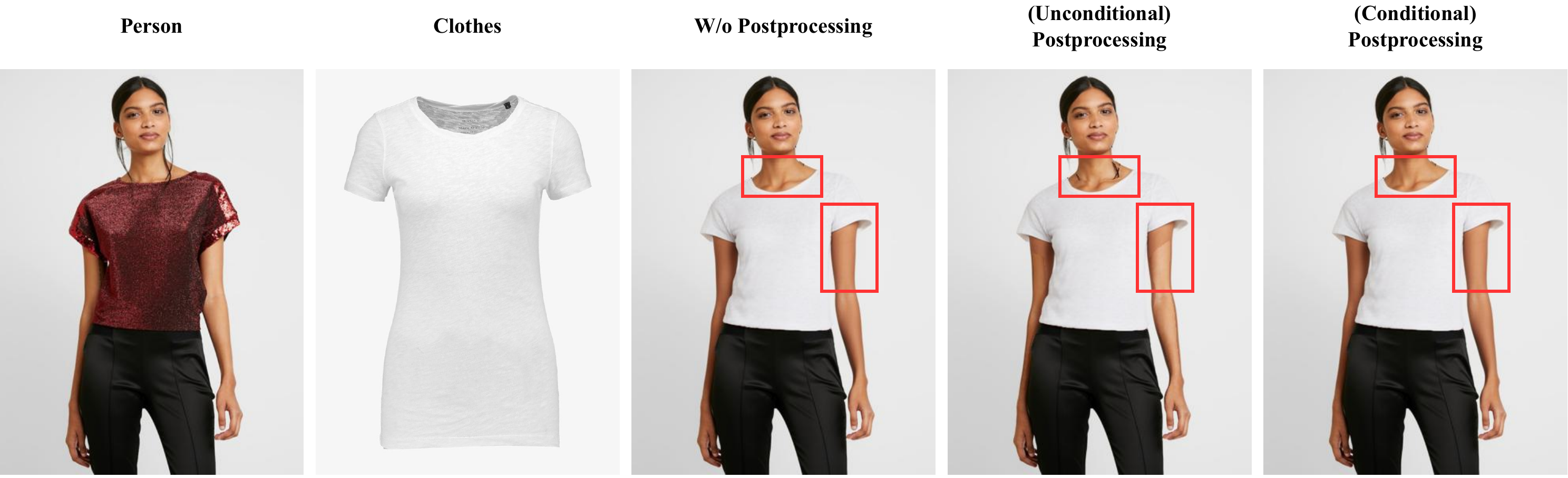}}
\caption{Visualization for conditional and unconditional post-processing block. If we use unconditional post-processing for the second one, the final output might get the artifact feature - red rectangle regions.}
\label{fig: postprocessing-errors}
\end{figure}
\squeezeup
\section{Reason for using conditional post-processing.} As mentioned above, we apply the second post-processing block with the condition of overlapping rate. To explain this, we illustrated the result if we used both post-processing blocks unconditionally, the final output can not handle the artifact since the model can not guarantee to generate the same as the real image - as depicted in Figure \ref{fig: postprocessing-errors}. 
\squeezeup
\section{Implemention Details.}
For the two main modules of our model, the warping module and the try-on module, we conduct separate training. The warping network is trained for 250 epochs using the Adam optimizer \cite{Adam} with a learning rate of \(5 \times 10^{-6}\). Regarding the try-on module, we train it for 500 epochs with the Adam optimizer and a learning rate of \(5 \times 10^{-5}\). It is noteworthy that for each resolution, we train different configurations for both warping and try-on module networks, and the hardware configuration for training is specified in Table \ref{tab:Network_Conf}, Table \ref{tab:Warping_HP}, and Table \ref{tab:Tryon_HP}.

\begin{figure}[t]
\centering
\scalebox{0.8}{\includegraphics[
  width=12cm,
  height=6cm,
  keepaspectratio,
]{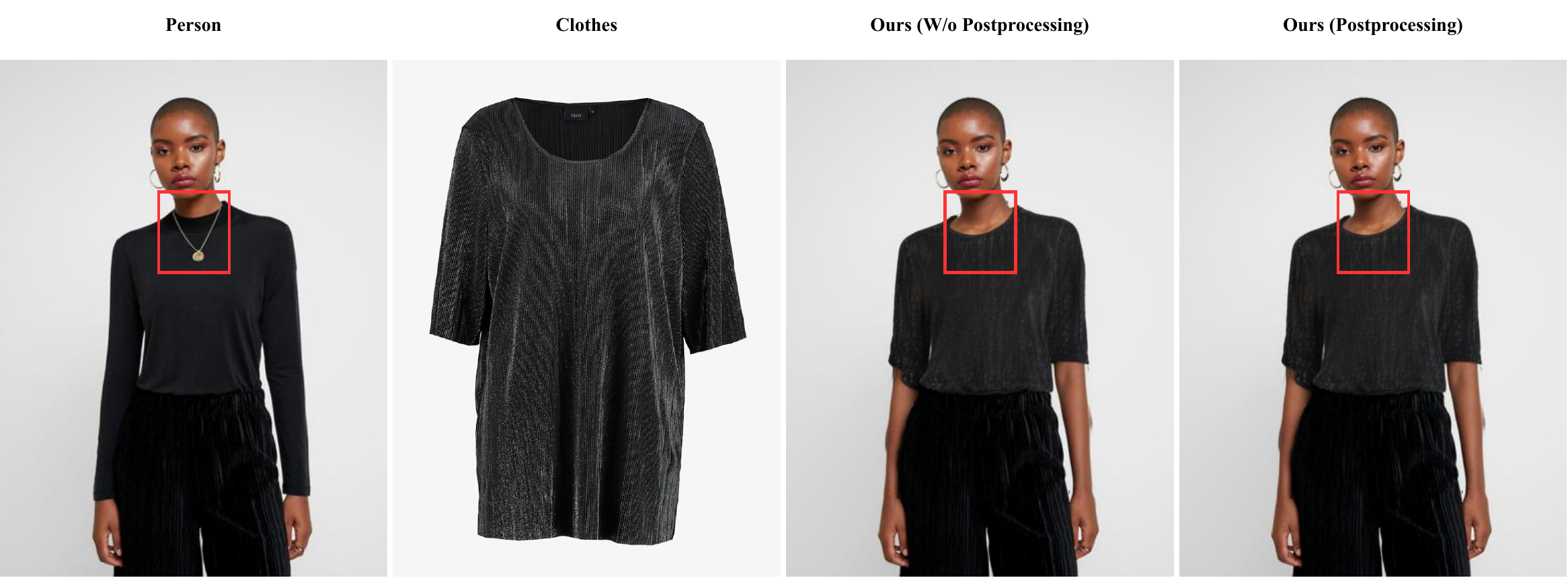}}
\caption{Identity remaining failure.}
\label{fig: Identity remaining failure}
\end{figure}
\squeezeup
\section{Discussion.}
\subsection{Conditional Post-processing Limitation.} 
As mentioned in the ablation study, conditional mask-aware post-processing directly affects the final model performance both qualitatively and quantitatively. Because the overlapping rate is considered the major condition, it greatly relies on the segmentation output of the body part from both the pre-processing human parsing step \cite{HumanParsing} and our warping module performance. Consequently, the effectiveness of the conditional mask-aware post-processing is inherently tied to the performance of these human parsing predictions. It encounters limitations when dealing with identity information that is situated on the clothing before the virtual try-on process. For example, the identity information lies on the clothing before trying the cloth on, as depicted in Fig. \ref{fig: Identity remaining failure}. It highlights the need for advanced segmentation and parsing techniques that can more precisely delineate between different types of identity information, ensuring that such details are faithfully preserved and accurately represented in the final output. This area presents an opportunity for further research and development, aiming to enhance the model's ability to handle a broader range of identity information with greater fidelity.   
\squeezeup
\subsection{Try-on Module Limitation.} 
As mentioned, our try-on module only focuses on generating the missing part in the image, fine-tooling the items of clothing, and all the preserved parts remain. Theoretically, this approach only works perfectly under circumstances where the segmentation preprocessing and the warping-based segmentation prediction are executed flawlessly. Meanwhile, that perfection never happens, hence, our diffusion-based module performance is again limited under these segmentation performances. Any inaccuracies or imperfections in the segmentation predictions can lead to errors in the final output, such as misaligned clothing, incorrect texture synthesis, or artifacts in the preserved regions. This limitation underscores the importance of continued advancements in segmentation techniques and the development of more robust models capable of handling imperfections in segmentation predictions.
\squeezeup
\subsection{Explicit Warping Module Limitation.} 
Another limitation of our approach lines on our explicit warping module procedure. This procedure will always generate that appears to have the same fit as the initial clothing the person is wearing. It is because of the use of human parsing as the condition where the preserved mask is used for the input condition.  

\begin{table}[h]
\centering
\caption{Network Configurations}
\scalebox{0.8}{
\begin{tblr}{
  width = \linewidth,
  colspec = {Q[87]Q[263]Q[190]Q[190]Q[204]},
  cells = {c},
  hline{1-2,5,12} = {-}{},
}
\textbf{Module} &  \textbf{Parameters} & \textbf{VITON-HD (256x192)} & \textbf{VITON-HD (512x384)} & \textbf{DressCode (512x384)} \\
Warping         & Number
  of FPN                  & 5                          & 6                          & 6                            \\
                & Cross-attention
  resolutions    & {[}64,32,16,8]             & {[}64,32,16,8]             & {[}64,32,16,8]               \\
                & Cross-attention
  dropout        & 0.2                        & 0.2                        & 0.2                          \\
Try-on           & Base
  channels                  & 256                        & 128                        & 128                          \\
                & Channels
  multiplier per scale  & {[}1, 2, 2, 2, 4]          & {[}1, 1, 2, 2, 4]          & {[}1, 1, 2, 2, 4]            \\
                & Attention
  dropout              & 0.1                        & 0.1                        & 0.1                          \\
                & Attention
  resolutions          & {[}64,32]                  & {[}64,32,16]               & {[}64,32,16]                 \\
                & Number
  of ResBlock
  per scale & 2                          & 2                          & 2                            \\
                & CLIP
  image encoder version     & ViT-B/32                   & ViT-B/32                   & ViT-B/32                     \\
                & Noise
  level                    & 5                          & 5                          & 5                            
\end{tblr}}
\label{tab:Network_Conf}
\end{table}

\begin{table}[h]
\centering
\caption{Warping Module Hyper-parameter Configurations}
\scalebox{0.8}{
\begin{tblr}{
  width = \linewidth,
  colspec = {Q[200]Q[204]Q[204]Q[219]},
  column{even} = {c},
  column{3} = {c},
  cell{9}{2} = {c=3}{0.627\linewidth},
  cell{10}{2} = {c=3}{0.627\linewidth},
  cell{11}{2} = {c=3}{0.627\linewidth},
  cell{12}{2} = {c=3}{0.627\linewidth},
  cell{13}{2} = {c=3}{0.627\linewidth},
  cell{14}{2} = {c=3}{0.627\linewidth},
  hline{1-2,9,15} = {-}{},
}
  & \textbf{VITON-HD (256x192)} & \textbf{VITON-HD (512x384)} & \textbf{DressCode (512x384)} \\
\(lr^w_G\)                                & \(5 \times e^{-6}\)       & \(5 \times e^{-6}\)         & \(5 \times e^{-6}\)                        \\
\(lr^w_D\)                                & \(5 \times e^{-6}\)        & \(5 \times e^{-6}\)      & \(5 \times e^{-6}\)                        \\
Adam (\({\beta}_1 - {\beta}_2\)) & 0.5 - 0.999                & 0.5 - 0.999                & 0.5 - 0.999                  \\
Batch size                         & 8                          & 4                          & 4                            \\
Number of epochs                   & 500                        & 250                        & 250                          \\
EMA                                  & None                       & None                       & None                         \\
Number of GPUs and GPUs            & 1 RTX 4090 (24GB)          & 1 A100 Tesla (40GB)        & 1 A100 Tesla (40GB)          \\
\({\alpha}^w_{per}\)                 & 0.2                        &                            &                              \\
\({\alpha}_{ce}\)                    & 3                          &                            &                              \\
\({\alpha}^w_m\)                     & 0.3                        &                            &                              \\
\({\alpha}^w_{adv}\)                 & 0.1                        &                            &                              \\
\({\alpha}_{TV}\)                    & 0.1                        &                            &                              \\
\({\alpha}_{sec}\)                   & 6                          &                            &                              
\end{tblr}
}
\label{tab:Warping_HP}
\end{table}

\begin{table}[h]
\centering
\caption{Try-on Module Hyper-parameter Configurations}
\scalebox{0.8}{
\begin{tblr}{
  width = \linewidth,
  colspec = {Q[200]Q[204]Q[204]Q[219]},
  column{even} = {c},
  column{3} = {c},
  cell{9}{2} = {c=3}{0.627\linewidth},
  cell{10}{2} = {c=3}{0.627\linewidth},
  cell{11}{2} = {c=3}{0.627\linewidth},
  hline{1-2,9,12} = {-}{},
}
                                     & \textbf{VITON-HD (256x192)} & \textbf{VITON-HD (512x384)} & \textbf{DressCode (512x384)} \\
\(lr^{tryon}_G\)                                & \(5 \times e^{-5}\)                      & \(5 \times e^{-5}\)         & \(5 \times e^{-5}\)          \\
\(lr^{tryon}_D\)                                & \(5 \times e^{-5}\)                      & \(5 \times e^{-5}\)        & \(5 \times e^{-5}\)            \\
Adam (\({\beta}_1 - {\beta}_2\)) & 0.9 - 0.999                & 0.9 - 0.999                & 0.9 - 0.999                  \\
Batch size                         & 3                          & 3                          & 3                            \\
Number of epochs                   & 500                        & 500                        & 500                          \\
EMA                                  & 0.9999                       & 0.9999                       & 0.9999                         \\
Number of GPUs and GPUs            & 1 RTX 4090 (24GB)          & 1 A100 Tesla (40GB)        & 1 A100 Tesla (40GB)          \\
\({\alpha}^{tryon}_{per}\)                 & 1                          &                            &                              \\
\({\alpha}^{tryon}_{adv}\)                           & 0.1                        &                            &            \\   

\({\alpha}_{n}\)                           & 5                        &                            &   
\end{tblr}
}
\label{tab:Tryon_HP}
\end{table}

%
%

\end{document}